%% file: main.tex
\documentclass[conference]{IEEEtran}

\usepackage{cite}
\usepackage{amsmath,amssymb,amsfonts}
\usepackage{graphicx}
\usepackage{textcomp}
\usepackage{xcolor}
\usepackage{booktabs}
\usepackage{bm}
\usepackage{pifont}
\usepackage{url}
\usepackage{comment}
\usepackage{enumitem}
\usepackage{wrapfig}
\usepackage{tikz}
\usetikzlibrary{arrows.meta,positioning,fit,calc}
\usepackage[caption=false,font=footnotesize]{subfig}

\usepackage[linesnumbered,ruled]{algorithm2e}
\newcommand{\cmark}{\ding{51}} 
\newcommand{\xmark}{\ding{55}} 
\newcommand{\ls}[1]{{\sf\color{orange}[LS: #1]}}

\def\BibTeX{{\rm B\kern-.05em{\sc i\kern-.025em b}\kern-.08em
    T\kern-.1667em\lower.7ex\hbox{E}\kern-.125emX}}

\begin{document}

\title{Beyond Non-IID: Learner--Client Distribution Mismatch in Federated Learning}

\author{
\IEEEauthorblockN{Yiming Xie, Lili Su, and Ningfang Mi}
\IEEEauthorblockA{
Department of Electrical and Computer Engineering\\
Northeastern University, Boston, MA, USA\\
Email: xie.yimi@northeastern.edu; l.su@northeastern.edu; n.mi@northeastern.edu
}
}

\maketitle

\input{00_abstract.tex}

\input{01_introduction.tex}
\input{02_related.tex}

\input{03_motivation.tex}

\input{04_method.tex}
\input{05_experiment.tex}
\input{06_conclusion.tex}

\bibliographystyle{IEEEtran}
\bibliography{IEEEcustom}

\end{document}

%% file: 00_abstract.tex
\begin{abstract}
Federated learning systems are increasingly deployed to facilitate collaborative model training across a heterogeneous client population. Existing practice mostly implicitly assumes that the aggregated client data distribution is representative of the learner's target distribution or that learning from all available clients is uniformly beneficial for the learner distribution. 
However, such an assumption often does not hold in reality. 
Traditional client selection strategies in FL literature largely overlook such misalignment, while  
most existing work on multi-source transfer learning either requires direct access to local data or uses one-shot model/feature aggregation. 
In this paper, we take the initiative to understand and mitigate the impacts of such learner–client population misalignment. In particular, we consider the practical setting where the learner keeps a small proxy dataset.   
We observe that client contributions vary significantly across training rounds, and traditional technology is insufficient to identify beneficial sources under multi-source transfer diversity. 
Then, we propose a dynamic, influence-aware client selection framework that estimates each client’s potential utility 
to the learner’s optimization objective using proxy influence signals 
on a learner-specific proxy set. 
Via using leave-one-out evaluations, we 
prioritize the most informative sources of knowledge while controlling the negative impacts of statistical noise and data heterogeneity. 
Experiments on CIFAR-10 under heterogeneous data partitions demonstrate that our approach consistently outperforms static and dynamic baselines, achieving faster convergence and higher accuracy.

\end{abstract}

%% file: 01_introduction.tex
\section{Introduction}
\label{sec: intro}

Federated learning (FL) enables collaborative model training across multiple data owners without requiring direct access to local data and has emerged as a widely adopted paradigm for learning from distributed and heterogeneous sources \cite{mcmahan2017fedavg, kairouz2021advances}. In classical FL, the objective is to optimize a global model that minimizes the average loss across clients, i.e.,

\begin{align}
\label{eq: population-wide opt}
\frac{1}{N}\sum_{i=1}^N F(\mathcal{D}_i, w),    
\end{align}
where $w$ denotes the model parameters, $N$ is the number of clients, $\mathcal{D}_i$ is the local dataset of client $i$, and $F(\mathcal{D}_i, w)$ is the corresponding local loss evaluated at $w$. 
This formulation implicitly treats all clients symmetrically with respect to the learner's goal. 
In many practical deployments, however, {\em the goal should not be population-wide optimization, but rather to improve the performance of a \emph{specific learner} under a target distribution of interest.} 

\begin{figure}[tb]
    \centering
    \includegraphics[width=\linewidth]{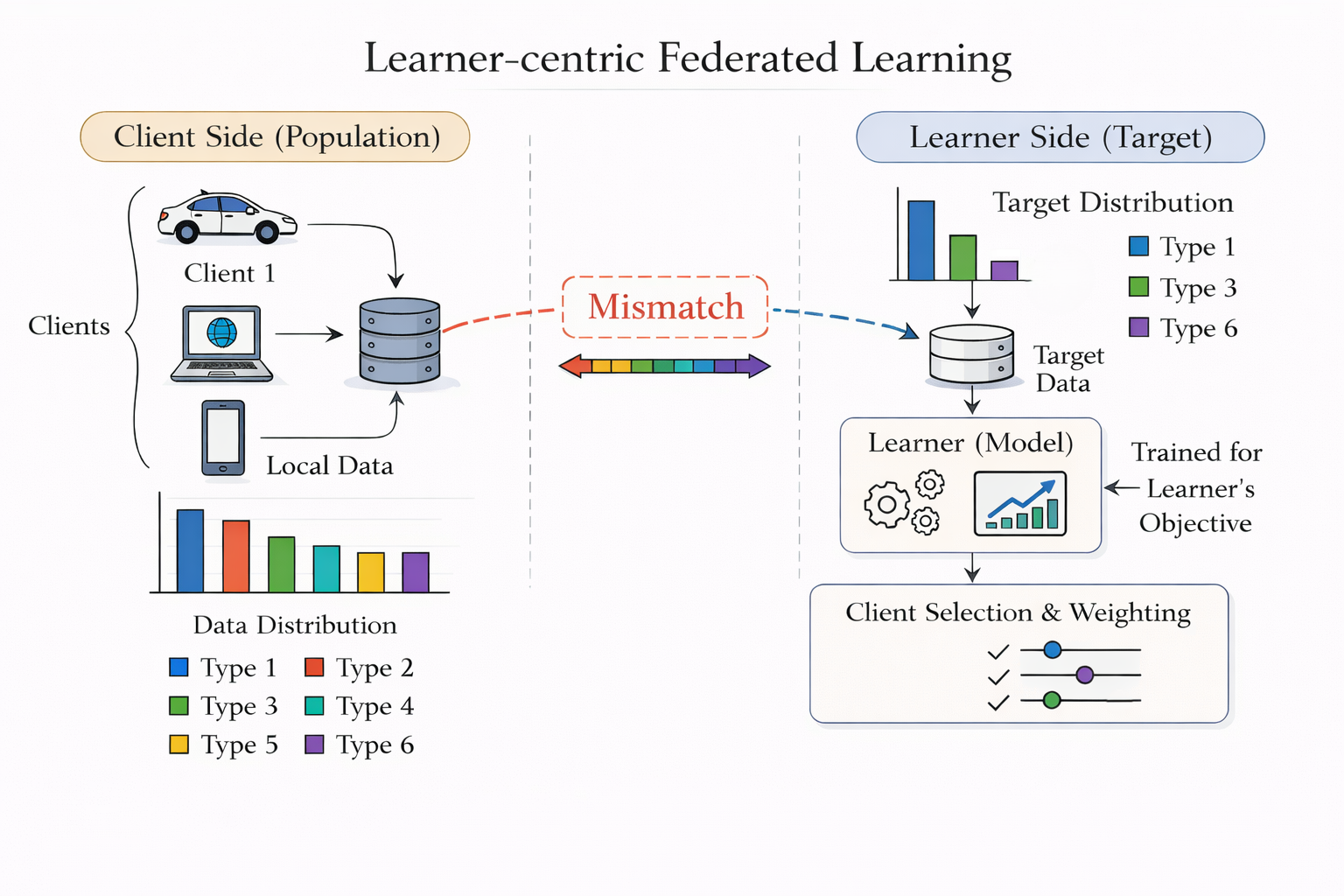}
    \vspace{-4em}
    \caption{Mismatch of client population data and target distribution. The left panel illustrates model training, where a learner receives updates from heterogeneous client devices. The empirical client population induces a training data distribution shown at the bottom left. The right panel illustrates the learner side, where the trained model is deployed to target users or applications, whose data distribution can differ substantially from the training distribution.}
    \label{fig:mismatch}
\end{figure}
This is because, as illustrated in Fig.~\ref{fig:mismatch}, the data distribution of participating clients may not align well with the distribution encountered at deployment, leading to a nontrivial data mismatch. 
For example, in mobile and edge learning systems, opt-in participants may represent a narrow demographic configuration, whereas the deployed model must serve a broader and potentially different user base \cite{bonawitz2019towards, kairouz2021advances}. In such settings, learning from all available clients can dilute information relevant to the target population and even induce negative transfer. Hence, in this paper, we study learner-centric FL, where the learner seeks to optimize a target-specific objective with the aid of a small proxy dataset by selectively leveraging a population of clients whose data distributions may not align with the target. A formal problem formulation can be found in Section \ref{subsec:system_model}. 
%
%
However, most existing client selection methods in FL 
%
select clients to improve system efficiency or to accelerate convergence \cite{8761315,bonawitz2019towards,9084352}.   
These approaches are not immediately applicable to our setting because they are designed to optimize a population-level objective, overlooking 
the learner-client population misalignment.  
%

Learner-centric FL can be viewed as a form of multi-source transfer learning, where a target learner aims to improve performance using knowledge from a heterogeneous set of client sources. 
In this context, both the \emph{quality} (distributional alignment) and the \emph{quantity} (amount of data or influence) of client contributions critically affect transfer performance \cite{ben2010theory,guo2018selective,pan2010survey,tong2021mathematical,wu2024h}; see Section \ref{sec:related} for details.    
Most of the existing transfer learning methods assume centralized access to source data or pretrained models. 
Recently, Wu et al.~\cite{wu2024h} propose the H-Ensemble method, which is a multi-source–free transfer approach that does not require raw data from the sources to be revealed to the target task. However, this method 
overlooks the potential for knowledge refinement in adapting the features.

To address these limitations, we propose a Dynamic Influence-Aware Control for Knowledge Transfer (DIC-KT) framework for learner-centric federated learning. 
Our approach estimates the marginal contribution of each client to the learner’s objective using a small learner-specific proxy dataset, without requiring access to raw client data. 
By using leave-one-out evaluations,  we prioritize the most
informative sources of knowledge while controlling the negative
impacts of statistical noise and data heterogeneity. 

Our contributions can be summarized as follows: 
\begin{itemize}
\item We formalize \emph{learner–client distribution mismatch} as a learner-centric multi-source transfer learning problem within federated learning.
\item We analyze the limitations of existing client selection strategies that rely on static or traditional dynamic selection methods under distribution mismatch and limited learner data.
\item We propose an influence-aware client selection framework that dynamically estimates each client’s marginal contribution to the learner’s objective without requiring access to raw client data.
\item We demonstrate consistent empirical improvements across heterogeneous and misaligned data partitions, highlighting the importance of adaptive source selection for effective knowledge transfer.
\end{itemize}











%% file: 02_related.tex
\section{Related Work}
\label{sec:related}
\subsection{Federated Learning} 
In real-world FL, local datasets are often non-IID across clients, causing local SGD updates to drift in conflicting directions~\cite{mcmahan2017fedavg}. 
Extensive evidence shows that non-IID local datasets can cause unstable optimization, slower convergence, and degraded accuracy \cite{zhao2018federated,karimireddy2020scaffold,wang2020convergence}. 

To improve robustness under heterogeneity, many approaches modify the training dynamics while still optimizing a population-wide objective, implicitly assuming that the target data distribution reflects the aggregate data across all clients. 
Proximal regularization (e.g., FedProx) constrains local updates to remain close to the global model, thereby reducing drift \cite{li2020fedprox}. Control-variate methods, such as SCAFFOLD, reduce update bias by correcting local gradients using global information \cite{karimireddy2020scaffold}. 
Instead of training a global model, personalized FL (PFL) separates commonality and heterogeneity by learning client-specific parameters 
\cite{li2021ditto,fallah2020personalized,smith2017federated,li2021survey}. Although these methods substantially improve stability and personalization, they typically evaluate success using population-average performance across participating clients.

However, in learner-centric FL, the goal is not to optimize the average client objective, but to improve performance on a specific \emph{learner target distribution}, which may significantly mismatch 
the aggregate data across all clients. 
The above training-dynamics fixes \cite{li2020fedprox,karimireddy2020scaffold,li2021ditto,fallah2020personalized,smith2017federated,li2021survey} alone do not resolve this mismatch because they do not explicitly identify which client updates are most beneficial for the learner. 
To provide a focus, in this paper, we study training a global model.

\subsection{Multi-Source Transfer Learning} 

Multi-source transfer learning studies how to leverage multiple auxiliary sources (datasets or models) to improve performance on a target task \cite{pan2010survey}.  
A central challenge is that different sources may vary substantially in relevance and quality, and transferring knowledge indiscriminately from all available sources can lead to \emph{negative transfer}; that is, compared with standalone training, target performance may deteriorate due to poorly aligned sources \cite{ben2010theory,guo2018selective}.

To address this issue, a large body of work investigates source selection and weighting strategies that prioritize sources most beneficial to the target task. 
Early theoretical work on multi-source domain adaptation shows that the target risk depends on the divergence between source and target distributions \cite{mansour2009domain}. 
Practical approaches estimate source usefulness through distributional similarity, representation alignment, or validation performance across domains \cite{gong2013connecting,zhao2018adversarial,peng2019moment}. 
These methods aim to identify source domains whose knowledge can transfer effectively while avoiding sources that introduce bias.

Recent studies further show that both source relevance and data quantity are critical for effective knowledge transfer—transferring “too much” knowledge from a relevant source can be highly suboptimal \cite{guo2018selective,tong2021mathematical,wu2024h}. 
For example, Tong et al.~propose to measure the transferability of different sources in terms of the optimal coefficients of a linear combination of source models, and characterize this notion of transferability in terms of sample size, model complexity, and the $\chi^2$-distance between source and target tasks~\cite{tong2021mathematical}. 

Most multi-source transfer learning typically assumes centralized access to source datasets and often relies on a one-shot assessment of source utility before performing transfer \cite{tong2021mathematical,wu2024h,sun2011two,zhao2018adversarial}. 
%
Recently, Wu et al.~\cite{wu2024h} propose the H-Ensemble method, which is a multi-source–free transfer approach that does not require raw data from the sources to be revealed to the target task. However, this method fuses the feature extractors learned from individual sources and overlooks the potential for knowledge refinement in adapting the features.  
{\color{blue}
}

\subsection{Adaptive Client Selection in FL}

Recent work in federated learning increasingly investigates adaptive client selection, where participation decisions are dynamically updated across training rounds. 
System-oriented strategies focus on improving training efficiency by prioritizing faster or more reliable clients to mitigate stragglers, reduce communication delays, or account for device availability \cite{8761315,bonawitz2019towards,9084352}. 
While effective for improving system throughput, these approaches are largely agnostic to the data characteristics of participating clients and therefore do not explicitly consider the learning impact of client data.

A second line of work focuses on data-aware or learning-driven client selection, particularly under heterogeneous data distributions. 
These methods attempt to select clients based on signals derived from training dynamics, such as gradient or update magnitudes \cite{wang2020optimizing,chai2020tifl}, historical loss reduction \cite{8761315}, or estimated client contribution \cite{wang2020optimizing,273723}.
Such heuristics aim to accelerate convergence under non-IID data by prioritizing clients that produce larger optimization progress or more informative updates. 
However, these signals primarily reflect optimization behavior under the global training objective and do not explicitly capture the relationship between client data distributions and the learner’s target task.

More direct approaches attempt to estimate client contribution through validation-based marginal effects or influence-style approximations. 
For example, FedInfluence estimates the effect of each client on validation loss using influence-function-inspired approximations of leave-one-out retraining \cite{Xue2020TowardUT}. 
Similarly, data valuation methods measure the utility of client participation by evaluating how validation performance changes when a client is included or excluded from aggregation \cite{wang2020principled}. 
While these approaches move closer to utility-aware client participation, they incur nontrivial computational overhead, such as influence-function approximations or repeated validation evaluations across candidate clients. 
More importantly, these methods evaluate contribution with respect to a global validation objective, implicitly assuming that improving population-level model performance is the primary goal.

In contrast, our work studies a learner-centric setting that requires client selection mechanisms to explicitly account for target–client distribution mismatch and to adaptively select clients based on their evolving potential contributions with respect to the model updates.

%% file: 03_motivation.tex
\section{Multi-Source Transfer Diversity}
\label{sec:multi_transfer_diversity}
In learner-centric federated learning, the goal is not to optimize a population-wide model, but to improve performance for a specific learner under its target distribution. In this setting, clients act as external knowledge sources whose utility may vary significantly due to differences in data relevance and transferability. 


\subsection{Challenges and Limitations of Existing Methods}

Table~\ref{tab:method_comparison} summarizes representative client and model selection approaches in federated and multi-source transfer learning. Most existing methods rely on either static similarity measures (e.g., distribution similarity) or proxy optimization signals such as local training loss. Although these strategies can improve training efficiency in conventional FL settings, they do not explicitly account for learner-centric utility under target-client distribution mismatch. In particular, they treat client usefulness as largely static and therefore cannot adapt to the evolving contribution of each client during training. 

\begin{table*}[t]
\centering
\caption{
Comparison of representative client/model selection criteria in federated and multi-source transfer learning.
Methods are categorized as \emph{static} or \emph{dynamic} based on whether their selection signal
is updated across training rounds.
}
\label{tab:method_comparison}

\small
\setlength{\tabcolsep}{7pt}
\renewcommand{\arraystretch}{1.15}

\begin{tabular}{lccccc}
\toprule
Method 
& Multi-Source 
& Target Data Scarcity 
& Learner-Centric 
& Target Label 
& Static / Dynamic \\
\midrule
FedAvg \cite{mcmahan2017fedavg}
& \cmark & \xmark & \xmark & Supervised & Static \\

FL+HC \cite{briggs2020federated}
& \cmark & \xmark & \xmark & Supervised & Static \\

MCW \cite{liu2021model} 
& \xmark & \cmark & \xmark & Supervised & Static \\

LogME \cite{you2021logme} 
& \xmark & \cmark & \xmark & Unsupervised & Static \\

H-Ensemble \cite{wu2024h} 
& \cmark & \xmark & \xmark & Unsupervised & Static \\

OTQMS \cite{Zhang2025AHS} 
& \cmark & \cmark & \xmark & Unsupervised & Dynamic \\

Power of Choice \cite{alistarh2020power} 
& \cmark & \cmark & \xmark & Supervised & Dynamic \\

\textbf{DIC-KT (\emph{Ours})}
& \cmark & \cmark & \cmark & Supervised & Dynamic \\
\bottomrule
\end{tabular}
\end{table*}
\subsection{Empirical Insights under Multi-Source Transfer Diversity}
\label{subsec:empirical_insights}

Consequently, simply sampling clients uniformly or relying on static similarity measures can lead to inefficient training and unstable convergence for the learner. This limitation becomes particularly severe when client distributions are highly heterogeneous and their transfer utility evolves during training.
This challenge motivates the need for adaptive client selection mechanisms that directly evaluate the impact of client updates on the learner's objective. Therefore, we propose an influence-based client selection strategy that dynamically estimates each client’s contribution to the learner’s objective, prioritizing beneficial sources while mitigating the impact of weakly relevant or harmful updates.



To systematically study multi-source transfer diversity, we construct four CIFAR-10~\cite{Krizhevsky09learningmultiple} partition settings grouped into two categories: homogeneous and heterogeneous. In the homogeneous settings, all 
clients are approximately uniformly distributed over the ten CIFAR-10 labels, such that each client contains near-balanced samples from every label and shares identical support. Under this condition, diversity arises primarily from the mismatch between the learner’s target distribution and the uniformly distributed sources. 

\begin{figure*}[t]
    \centering

    \subfloat[Case 1]{%
        \includegraphics[width=0.24\textwidth]{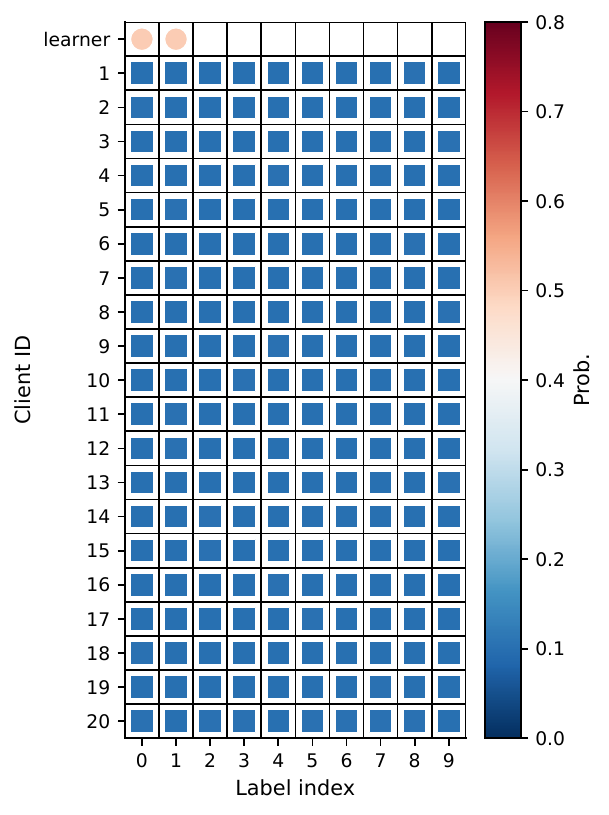}
    }
    \hfill
    \subfloat[Case 2]{%
        \includegraphics[width=0.24\textwidth]{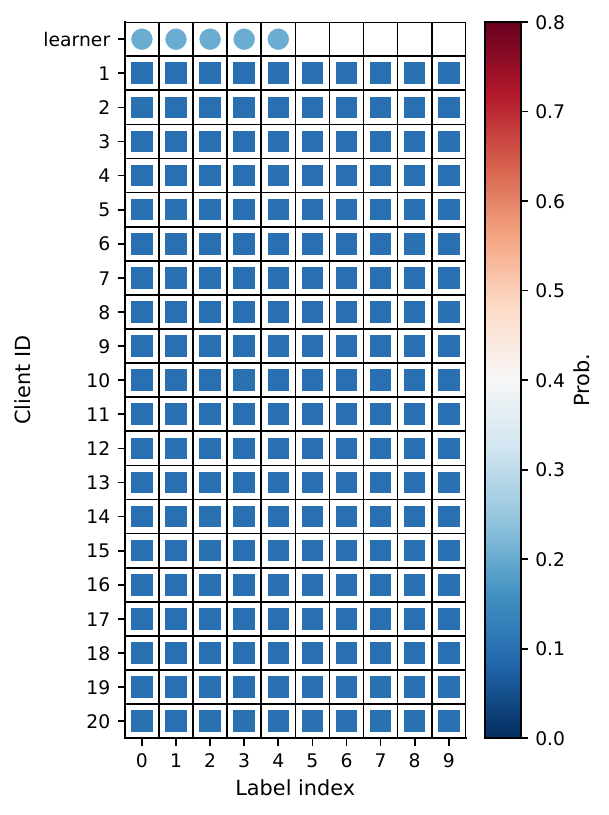}
    }
    \hfill
    \subfloat[Case 3]{%
        \includegraphics[width=0.24\textwidth]{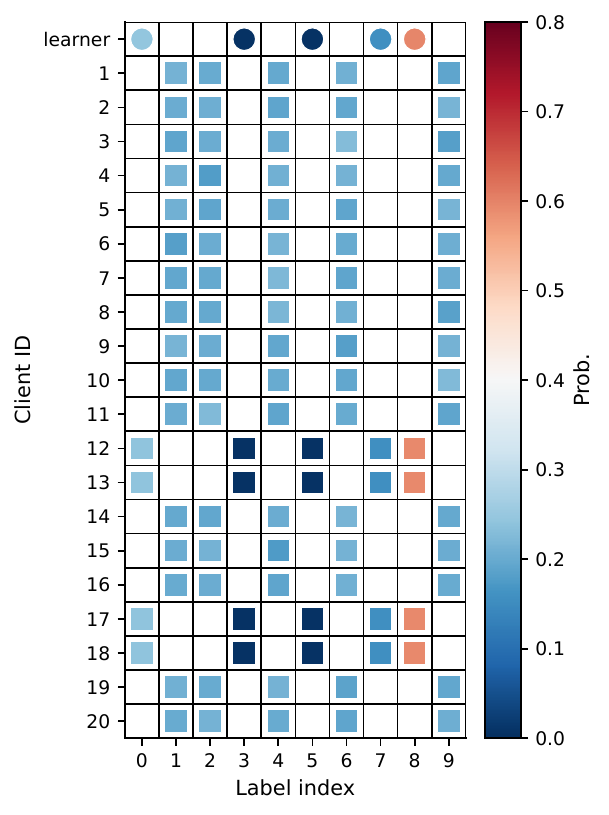}
    }
    \hfill
    \subfloat[Case 4]{%
        \includegraphics[width=0.24\textwidth]{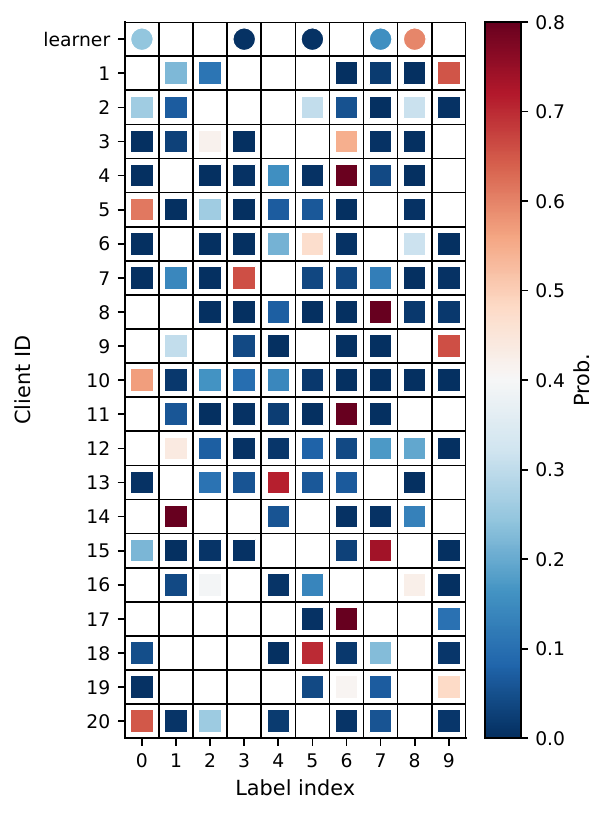}
    }

    \caption{Label distribution heatmaps across four experimental cases.
    The learner is shown with circular markers, while other clients use square markers.
    Color indicates class probability.
    }
    \label{fig:label_dist_all_cases}
\end{figure*}

\textbf{[Four Cases of Multi-source Transfer Diversity.]} As illustrated in Fig.~\ref{fig:label_dist_all_cases}, Case (1) restricts the learner to two target labels (e.g., $\{0,1\}$), while clients retain all ten classes, leading to strong target sparsity and limited overlap between the learner and sources. In Case (2), the learner's support expands to five labels (e.g., $\{0,1,2,3,4\}$), increasing the degree of overlap while maintaining homogeneous client distributions.

In contrast, the heterogeneous settings explicitly introduce inter-client diversity, i.e., Non-IID. In Case (3), the learner's target distribution is generated via a Dirichlet distribution with concentration parameter $\alpha=0.1$ over a fixed support set (e.g., $\{0,3,5,7,8\}$). Clients are then partitioned into two disjoint groups: {\em match} clients that contain only learner-support labels and approximately follow the learner's class proportions, and {\em exclude} clients that contain only complementary labels, creating maximal structural diversity across sources. In Case (4), all clients are generated using class-wise Dirichlet sampling with $\alpha=0.1$, producing skewed and non-identical label proportions across clients without strict support separation. 

\begin{figure*}[t]
\centering
\subfloat[Case 1]{%
    \includegraphics[width=0.24\textwidth]{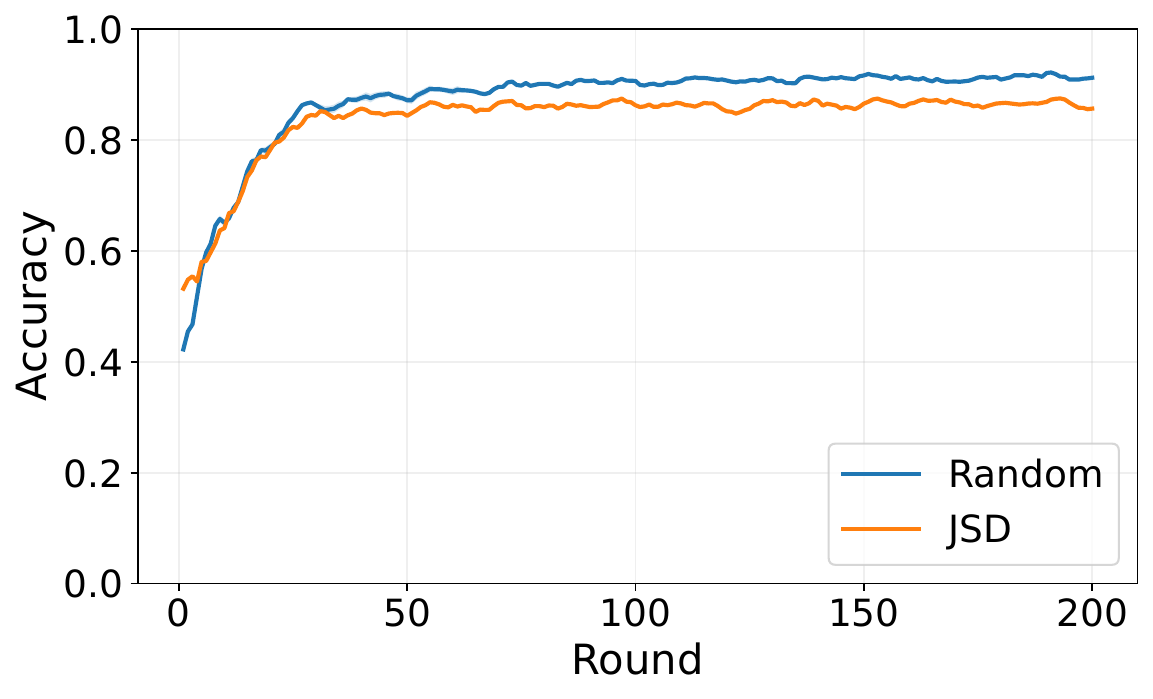}
}
\hfill
\subfloat[Case 2]{%
    \includegraphics[width=0.24\textwidth]{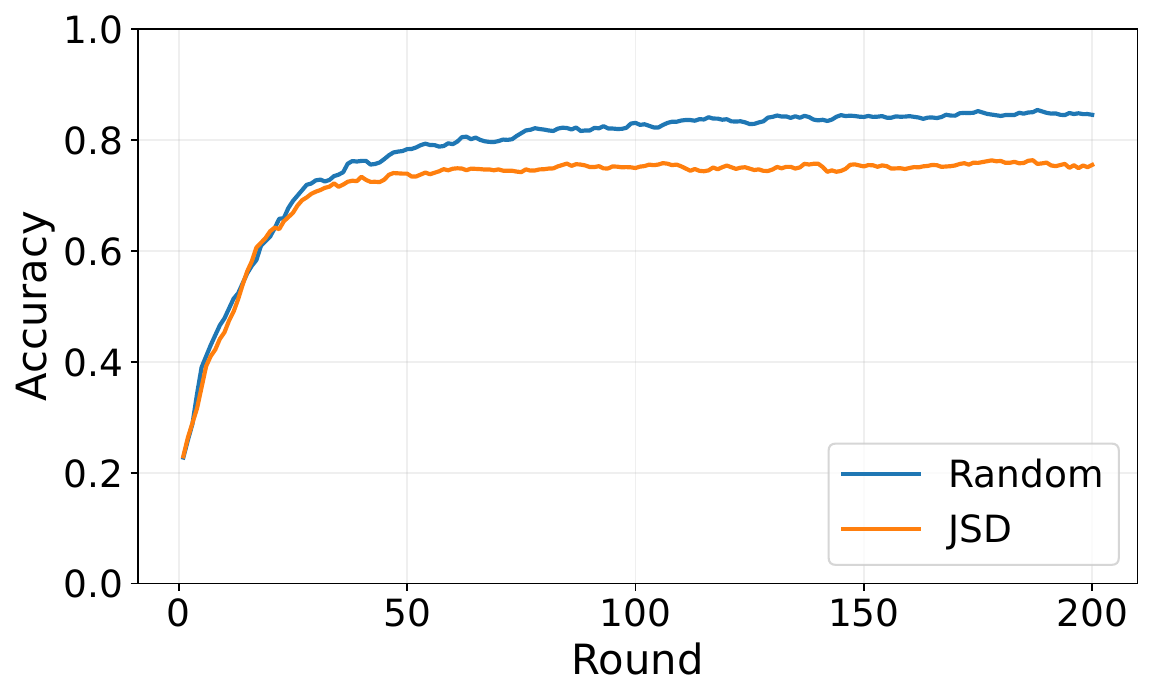}
}
\hfill
\subfloat[Case 3]{%
    \includegraphics[width=0.24\textwidth]{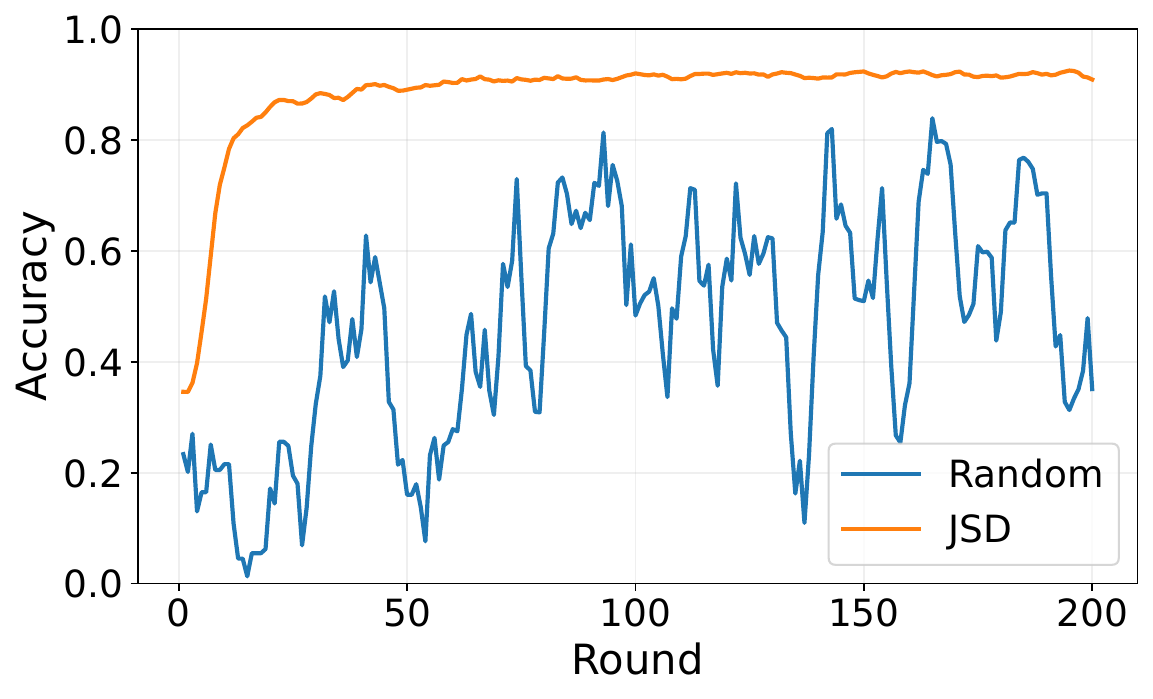}
}
\hfill
\subfloat[Case 4]{%
    \includegraphics[width=0.24\textwidth]{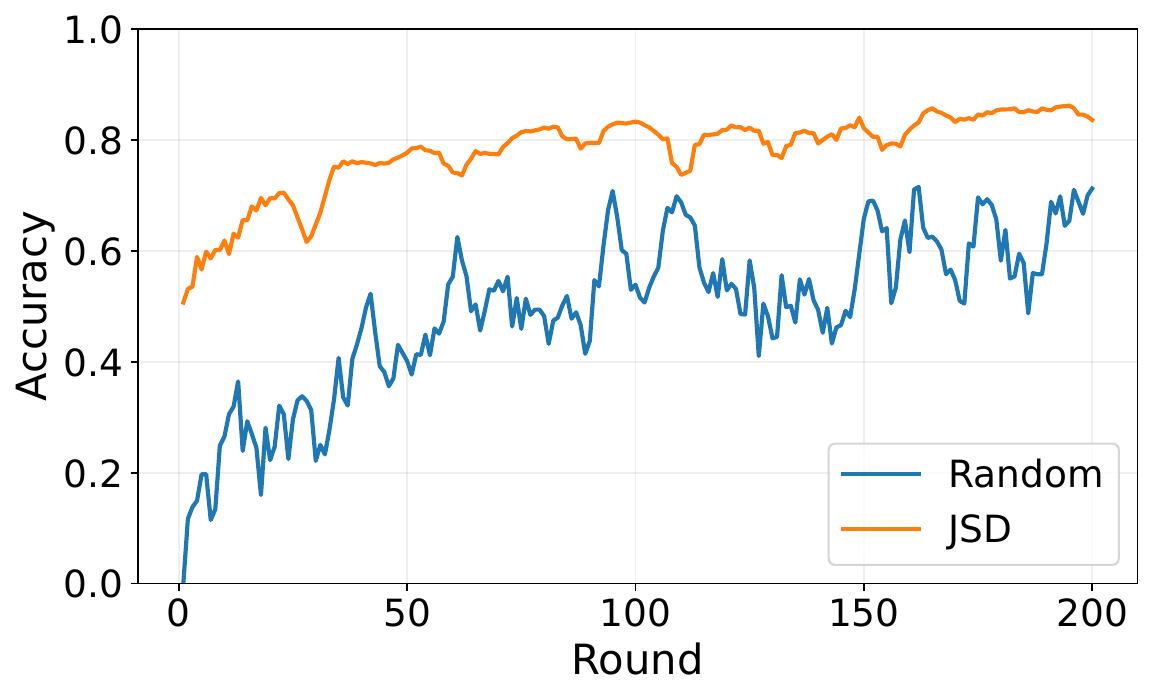}
}

\caption{Runtime accuracy across the four cases.}
\label{fig:runtime_acc_all}
\end{figure*}

\begin{figure*}[t]
\centering
\subfloat[Case 1]{%
    \includegraphics[width=0.24\textwidth,height=2.5cm]{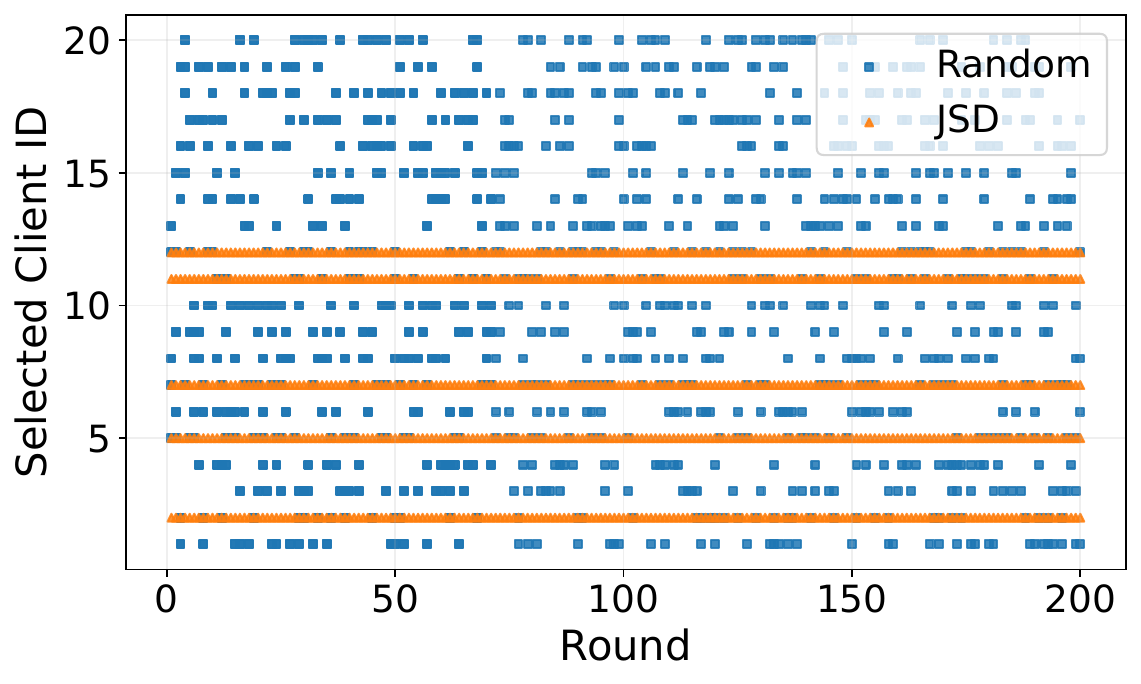}
}
\hfill
\subfloat[Case 2]{%
    \includegraphics[width=0.24\textwidth,height=2.5cm]{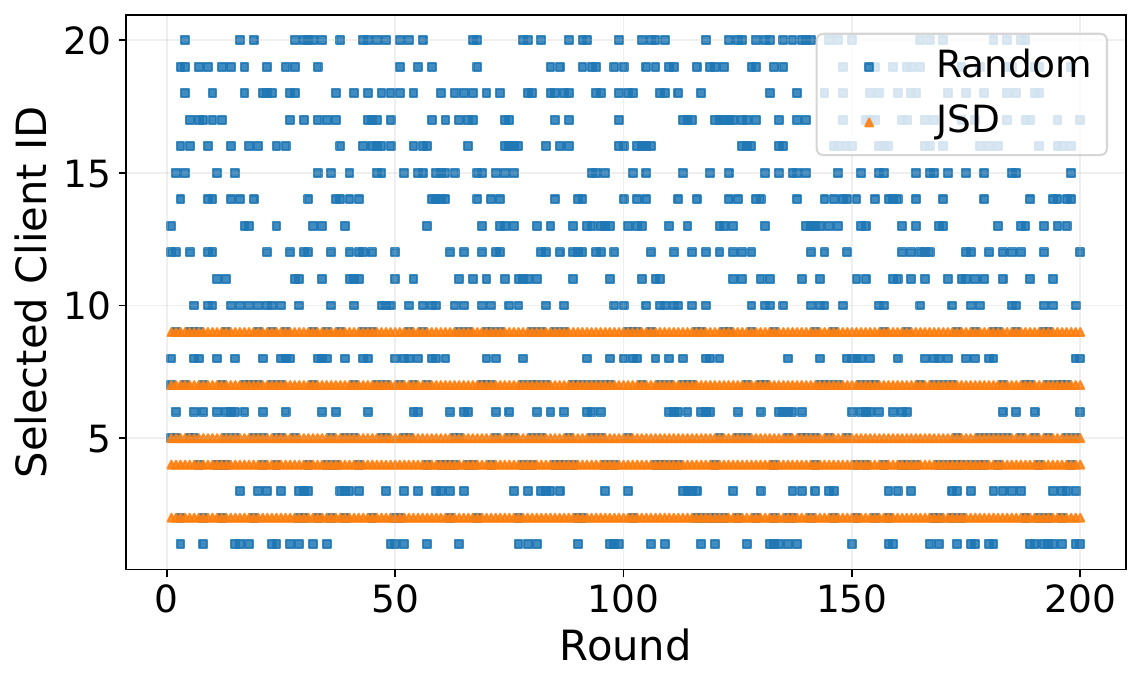}
}
\hfill
\subfloat[Case 3]{%
    \includegraphics[width=0.24\textwidth,height=2.5cm]{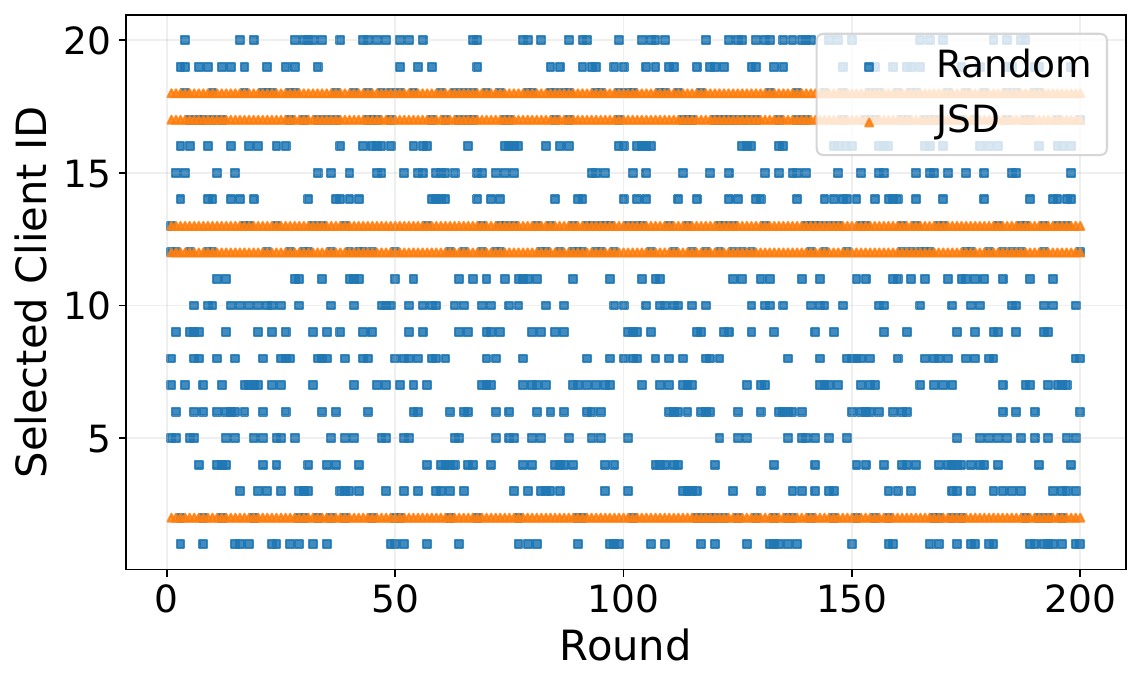}
}
\hfill
\subfloat[Case 4]{%
    \includegraphics[width=0.24\textwidth,height=2.5cm]{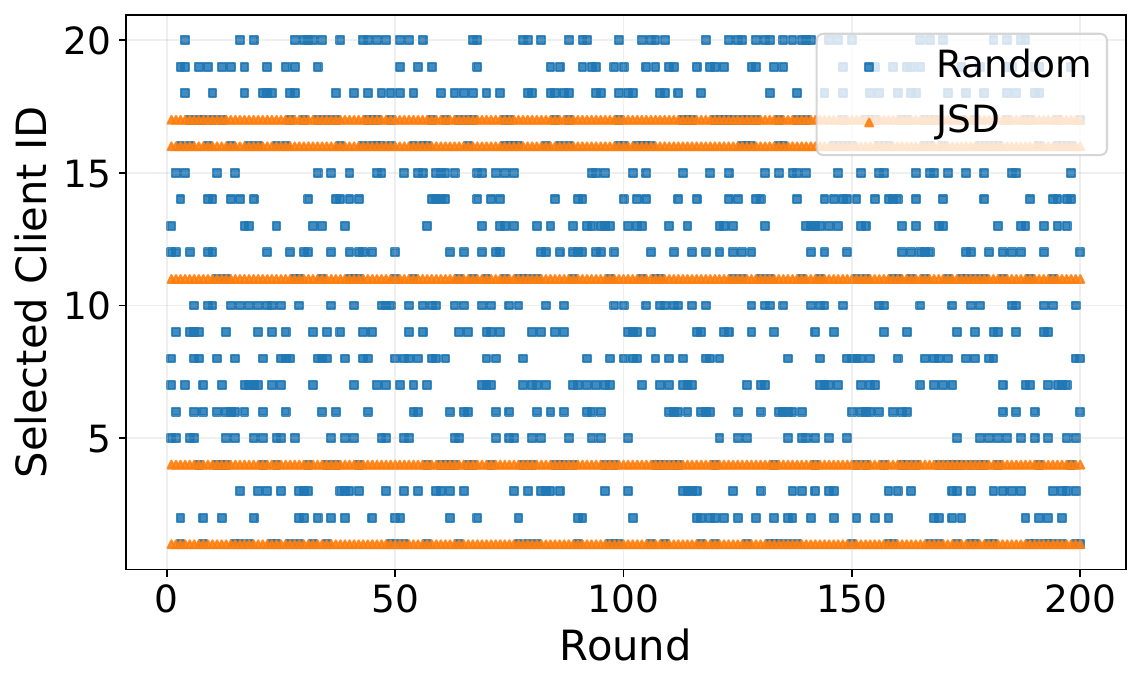}
}

\caption{Selected client IDs per round across the four cases.}
\label{fig:runtime_sel_all}
\end{figure*}

Together, these four cases progressively increase multi-source transfer diversity by varying both the learner's label support and the degree of inter-client heterogeneity, enabling systematic evaluation of alignment-aware client selection mechanisms under controlled homogeneous and heterogeneous data regimes.
Fig.~\ref{fig:runtime_acc_all} and Fig.~\ref{fig:runtime_sel_all} present
the runtime accuracy and the corresponding client selection trajectories under these four cases, revealing several key insights into how different selection strategies respond to diversity in multi-source transfer.

\noindent \textbf{[Homogeneous Source Cases.]} First, in the homogeneous settings (i.e., Cases~1 and~2), all clients exhibit approximately uniform label distributions across the ten CIFAR-10 classes, resulting in minimal inter-client distribution differences. Consequently, similarity-based metrics such as JSD~\cite{briggs2020federated} provide limited discrimination because client marginals are nearly identical. This behavior is reflected in the selection-ID plots (Fig.~\ref{fig:runtime_sel_all}(a)-(b)), where JSD repeatedly selects a relatively small subset of clients across rounds, whereas Random sampling naturally explores a broader range of clients. Such repetitive selection limits the diversity of incorporated updates and illustrates that distribution similarity provides little additional information when client distributions are already highly homogeneous.

\textbf{[Heterogeneous Source Cases.]} The heterogeneous settings (i.e., Cases~3 and~4) introduce explicit structural diversity among clients.
In Case~3, clients are split into two groups based on the learner's Dirichlet-generated label support: those matching the learner's labels (e.g., clients 12, 13, 17, and 18) and those containing only complementary labels.
This configuration produces a strong distributional contrast across clients, making JSD more informative for distinguishing aligned clients.
As shown in Fig.~\ref{fig:runtime_sel_all}(c), JSD increasingly focuses on the match-client subset over time, while Random sampling continues to select clients from both groups without considering their distributional relevance. This behavior demonstrates that distribution similarity becomes an effective selection signal when clear structural heterogeneity exists among clients.

Case~4 introduces a more general form of heterogeneity by applying Dirichlet sampling across all clients with $\alpha=0.1$.
Unlike Case~3, client label supports are no longer strictly separated. Clients share partially overlapping labels but exhibit highly skewed and diverse class proportions. Thus, heterogeneity arises primarily from differences in label proportions rather than explicit support separation.

In this case, JSD continues to adapt its selected client subset according to distribution similarity, while Random sampling maintains a more exploratory selection behavior. However, because client distributions partially overlap, similarity alone cannot fully characterize the usefulness of client updates. Consequently, distribution similarity provides weaker discrimination than in Case~3, highlighting the increased complexity of client selection under partially overlapping heterogeneous distributions.

\textbf{[Summary of Key Insights.]} The above results highlight an important characteristic of multi-source transfer diversity. In homogeneous settings, similarity-based methods, such as JSD, provide little discrimination because client distributions are nearly identical, resulting in repetitive client selection. As inter-client heterogeneity increases, JSD becomes increasingly effective at identifying distributionally aligned clients. However, when client distributions partially overlap, similarity alone is insufficient to fully characterize the utility of client updates. These observations demonstrate both the strengths and limitations of similarity-based client selection under different degrees of transfer diversity and motivate the development of more learner-aware client selection strategies. 

%% file: 04_method.tex
\section{Federated Learning with Influence Functions}

\label{sec:influence}



We propose a learner-centric client selection algorithm based on an influence function (DIC-KT), formally described in Algorithm~\ref{alg:lc-influence}. In the following, we provide a detailed description of the algorithm.

\subsection{Problem Setting}
\label{subsec:system_model}

We consider a federated learning system consisting of one \emph{learner} and
$C$ distributed clients indexed by $\mathcal{C}=\{1,2,\dots,C\}$.
Each 
client holds a private dataset generated from an underlying distribution over
the input--label space $\mathcal{X}\times\mathcal{Y}$, and raw data are never shared.

Unlike the distributed clients, the learner does not participate in local training and does not generate model updates. Instead, the learner serves as a central evaluation entity and maintains a small proxy dataset sampled from the target distribution. The learner uses this proxy dataset to assess the relevance and utility of client updates with respect to the learner's target task. This assumption is inspired by proxy-data-based evaluation mechanisms such as Zeno~\cite{xie2019zeno}, where a coordinating node maintains a small proxy dataset to assess the quality of distributed updates.

The learner is associated with a \emph{target distribution} $\mathcal{D}_0$ and observes a limited local dataset
\begin{equation}
\mathcal{S}_0 = \{(x_j,y_j)\}_{j=1}^{n_0} \sim \mathcal{D}_0,
\end{equation}
where $n_0$ is typically small. 
$\mathcal{S}_0$ serves as the learner's proxy dataset.

Each client $c \in \mathcal{C}$ holds its own private dataset
\begin{equation}
\mathcal{S}_c = \{(x_{c j},y_{c j})\}_{j=1}^{n_c} \sim \mathcal{D}_c,
\end{equation}
where the client distributions $\{\mathcal{D}_c\}_{c\in\mathcal{C}}$ may differ arbitrarily from one another
and from the learner’s target distribution $\mathcal{D}_0$.
In particular, the learner and clients need not be distributionally aligned, and similarity between
client data and the learner’s target task is not assumed.

We make no assumptions on the similarity, clustering structure, or parametric form of the client
distributions.
Client datasets may exhibit substantial statistical heterogeneity and may not admit a meaningful
partition into a small number of groups.
Moreover, the learner does not have access to client datasets or distributions and can only interact
with clients through model updates exchanged during federated training.


Unlike standard federated learning, the goal is not to minimize a global
population-wide objective.
Instead, the learner seeks to minimize its own target risk:
\begin{equation}
\min_{\theta\in\Theta}
\;\mathcal{L}(\mathcal{D}_0,\theta) \;\triangleq\;
\mathbb{E}_{(x,y)\sim\mathcal{D}_0}
\bigl[\ell(f_\theta(x),y)\bigr],
\label{eq:target_objective}
\end{equation}
where $\ell(\cdot,\cdot)$ is a task-specific loss function, and $f_\theta:\mathcal{X}\rightarrow\mathcal{Y}$ denotes a prediction model parameterized by $\theta \in \Theta \subseteq \mathbb{R}^d$.
Since $\mathcal{D}_0$ is unknown to all participants, the learner relies on its local proxy
dataset $\mathcal{S}_0$ 
to serve as a proxy for Eq.~\eqref{eq:target_objective}.

Minimizing the aggregate client objective
$\sum_{c\in\mathcal{C}}\mathcal{L}(\mathcal{D}_c,\theta)$
does not generally minimize $\mathcal{L}(\mathcal{D}_0,\theta)$ when client
distributions are mismatched with the learner’s target distribution.
This mismatch is particularly harmful when $|\mathcal{S}_0|$ is small.

\subsection{Influence-Based Client Contribution Estimation}
\label{subsec:influence}
%
%
%
In learner-centric federated learning, the value of a client is determined by whether its update helps reduce the learner's underlying generalization loss $\mathcal{L}(\mathcal{D}_0,\theta)$, which is not observable to every client and even the learner itself. 
To obtain a round-specific estimate of client utility without inspecting raw data, we use leave-one-out (LOO) \cite{Vehtari2015PracticalBM} evaluation on the learner proxy set. LOO provides a good candidate to measure the marginal effect of removing one client update from the current aggregation and therefore offers a practical signal for learner-centric client selection.

At communication round $r$, the learner broadcasts the current model parameters $\theta^{(r)}$
to a set of participating clients $\mathcal{C}^{(r)}\subseteq\mathcal{C}$.
Each client $c\in\mathcal{C}^{(r)}$ performs local optimization on its private dataset $\mathcal{S}_c$
for $E$ local epochs and returns an updated model $\theta_c^{(r)}$ along with its sample size $n_c$.
The learner collects the set of client models $\{(\theta_c^{(r)},n_c)\}_{c\in\mathcal{C}^{(r)}}$. 

As shown in Algorithm~\ref{alg:lc-influence}, lines~7--10, we first compute the standard sample-size weights
\begin{equation}
w_c^{(r)} \triangleq
\frac{n_c}{\sum_{j\in\mathcal{C}^{(r)}} n_j},
\qquad c\in\mathcal{C}^{(r)},
\label{eq:sample_weight}
\end{equation}
and define $w_c^{(r)}=0$ for $c\notin\mathcal{C}^{(r)}$. The corresponding sample-size-weighted aggregate model is
\begin{equation}
\bar{\theta}^{(r)}
\;\triangleq\;
\sum_{j\in\mathcal{C}^{(r)}} w_j^{(r)} \theta_j^{(r)}.
\label{eq:mean_model}
\end{equation}
Evaluating this aggregation on the learner proxy set $\mathcal{S}_0$ gives the learner empirical loss at round $r$:
\begin{equation}
\widehat{\mathcal{L}}_0^{(r)}
\;\triangleq\;
\widehat{\mathcal{L}}_0\!\left(\bar{\theta}^{(r)}\right).
\label{eq:full_loss}
\end{equation} 

%


To assess the marginal effect of client $c$, we construct the leave-one-out model that excludes client $c$:
\begin{equation}
\theta_{-c}^{(r)}
\;\triangleq\;
\frac{\bar{\theta}^{(r)} - w_c^{(r)}\,\theta_c^{(r)}}{1-w_c^{(r)}},
\qquad \text{for } w_c^{(r)} < 1,
\label{eq:loo_model}
\end{equation}
which is equivalent to sample-size-weighted averaging over clients in $\mathcal{C}^{(r)}\setminus\{c\}$ with renormalized weights. We then evaluate the leave-one-out model on the learner proxy set:
\begin{equation}
\widehat{\mathcal{L}}_{0,-c}^{(r)}
\;\triangleq\;
\widehat{\mathcal{L}}_0\!\left(\theta_{-c}^{(r)}\right).
\label{eq:loo_loss}
\end{equation}

Intuitively, if excluding client $c$ leads to a larger LOO loss ${\mathcal{L}}_{0,-c}^{(r)}$,
then client $c$ is more helpful to the learner at round $r$. 
Following the leave-one-out idea, we define the utility delta of client $c$ at round $r$ as
\begin{equation}
\Delta_c^{(r)}
\;\triangleq\;
\widehat{\mathcal{L}}_{0,-c}^{(r)} - \widehat{\mathcal{L}}_0^{(r)}.
\label{eq:delta_influence}
\end{equation}

To prioritize clients with positive learner-side utility, we convert the utility deltas into aggregation weights as shown in Algorithm~\ref{alg:lc-influence}, lines~17--20:
\begin{equation}
\begin{aligned}
\tilde{\lambda}_c^{(r)}
&\;\triangleq\;
\left(\max\{\Delta_c^{(r)},0\} + \epsilon^{1/|\Delta_c^{(r)}|}\right)^{\eta}, \\
\lambda_c^{(r)}
&\;\triangleq\;
\frac{\tilde{\lambda}_c^{(r)}}{\sum_{j\in\mathcal{C}^{(r)}} \tilde{\lambda}_j^{(r)}} .
\end{aligned}
\label{eq:influence_aggregation}
\end{equation}
where $\eta>0$ controls the sensitivity of the weighting rule and $\epsilon>0$ is a small stabilization constant. 
The utility delta $\Delta_c^{(r)}$ measures how the learner loss changes when client $c$ is removed from the aggregation. 
If $\Delta_c^{(r)}>0$, excluding client $c$ increases the learner loss, indicating that the client provides useful knowledge for the learner. 
In this case, the term $\max\{\Delta_c^{(r)},0\}$ remains positive, resulting in a larger value of $\tilde{\lambda}_c^{(r)}$ and therefore a larger aggregation weight.

If $\Delta_c^{(r)}\leq0$, removing the client does not increase the learner loss and may even reduce it, indicating that the client is redundant or potentially harmful to the learner objective. 
For such clients, the term $\max\{\Delta_c^{(r)},0\}$ becomes zero, and the weight is determined solely by the stabilization term $\epsilon^{1/|\Delta_c^{(r)}|}$. 
Because the exponent depends on the magnitude of $|\Delta_c^{(r)}|$, clients with larger negative utility produce smaller values of $\epsilon^{1/|\Delta_c^{(r)}|}$ and therefore receive smaller aggregation weights after normalization. 
This design ensures that harmful clients are not completely removed—preserving aggregation stability—while still being strongly down-weighted relative to helpful clients.

Finally, the exponent $\eta$ controls the strength of prioritization: larger values of $\eta$ amplify the difference between high-utility and low-utility clients, allowing the learner to focus more strongly on clients that provide greater improvements to its target objective.

\subsection{Knowledge Transfer Gain and Learner-Centric Utility}
\label{subsec:kt_gain}

We now formalize how the benefit of federated knowledge transfer is evaluated from the learner's perspective. 

\paragraph{No-transfer baseline.}
We first define a baseline learner model obtained without federated knowledge transfer.
Let $h_0^{\mathrm{self}}$ denote the model trained solely on the learner’s local dataset $\mathcal{S}_0$ using empirical risk minimization
\begin{equation}
\widehat{R}_0(h)
\triangleq
\frac{1}{|\mathcal{S}_0|}
\sum_{(x,y)\in\mathcal{S}_0}
\ell(h(x),y),
\label{eq:self_model}
\end{equation}
where the model parameters are optimized using stochastic gradient descent.

\paragraph{Transferred learner model.}
Let $\theta^{(T)}$ denote the learner model obtained after $T$ rounds of influence-aware federated training using the aggregation rule in~\eqref{eq:influence_aggregation}. In our experiments, $\theta^{(T)}$ corresponds to the final learner model produced by Algorithm~\ref{alg:lc-influence}.
This model integrates updates from multiple clients while prioritizing those that improve the learner's objective. 

\paragraph{Knowledge transfer gain.}
To quantify the benefit of incorporating external knowledge, we define the learner-centric knowledge transfer gain as
\begin{equation}
G
\;\triangleq\;
\mathcal{L}_0(\theta^{\mathrm{self}})
-
\mathcal{L}_0(\theta^{(T)}),
\label{eq:kt-gain}
\end{equation}
where $\mathcal{L}_0(\theta)$ denotes the learner population risk under the target distribution. 
A positive value $G>0$ indicates that federated training successfully transfers useful knowledge to the learner.

In practice, since the true distribution $\mathcal{D}_0$ is unknown, the risks in~\eqref{eq:kt-gain} are estimated using the learner's dataset $\mathcal{S}_0$. 

\paragraph{Learner-centric utility of clients.}
In heterogeneous environments, not all client updates are beneficial to the learner. 
Some clients may introduce biased or conflicting information that increases the learner loss. 
To address this challenge, our framework estimates each client's marginal utility through leave-one-out evaluation on the learner proxy set. 
Clients whose removal leads to a larger increase in learner loss receive higher influence weights, while harmful or redundant clients are down-weighted.

\begin{algorithm}[t]
\caption{Dynamic Influence-Aware Control for Knowledge Transfer (DIC-KT)}
\label{alg:lc-influence}
\DontPrintSemicolon
\footnotesize

\KwIn{Learner proxy set $\mathcal{S}_0$; clients $\mathcal{C}$ with local data $\{\mathcal{S}_c\}$; rounds $T$; local steps $E$; select $m$; $\eta>0$.}

\KwOut{Final learner model $\theta^{(T)}$.}

Initialize $\theta^{(0)}$\;

\For{$r=0,\dots,T-1$}{

    \tcp{Local training}
    Broadcast $\theta^{(r)}$ to available clients $\mathcal{C}^{(r)}$\;

    \ForEach{$c\in\mathcal{C}^{(r)}$}{
        $\theta_c^{(r)} \leftarrow \textsc{LocalTrain}(\theta^{(r)},\mathcal{S}_c,E)$\;
        upload $(\theta_c^{(r)},n_c)$\;
    }

    \tcp{Sample-size-weighted aggregation and learner loss}
    $N^{(r)} \leftarrow \sum_{j\in\mathcal{C}^{(r)}} n_j$\;
    
    $w_c^{(r)} \leftarrow n_c / N^{(r)}, \quad \forall c\in\mathcal{C}^{(r)}$\;
    
    $\bar{\theta}^{(r)} \leftarrow \sum_{j\in\mathcal{C}^{(r)}} w_j^{(r)} \theta_j^{(r)}$\;
    
    $\widehat{\mathcal{L}}_0^{(r)} \leftarrow \widehat{\mathcal{L}}_0(\bar{\theta}^{(r)})$\;

    \tcp{Leave-one-out utility estimation}
    \ForEach{$c\in\mathcal{C}^{(r)}$}{
        $\theta_{-c}^{(r)} \leftarrow (\bar{\theta}^{(r)}-w_c^{(r)}\theta_c^{(r)})/(1-w_c^{(r)})$\;
        
        $\widehat{\mathcal{L}}_{0,-c}^{(r)} \leftarrow \widehat{\mathcal{L}}_0(\theta_{-c}^{(r)})$\;
        
        $\Delta_c^{(r)} \leftarrow \widehat{\mathcal{L}}_{0,-c}^{(r)} - \widehat{\mathcal{L}}_0^{(r)}$\;
        

        $\tilde{\lambda}_c^{(r)} \leftarrow 
        \left(\max\{\Delta_c^{(r)},0\} + \epsilon^{1/|\Delta_c^{(r)}|}\right)^{\eta}$\;
    }

    \tcp{Normalize influence weights}
    $\lambda_c^{(r)} \leftarrow \tilde{\lambda}_c^{(r)} \big/ \sum_{j\in\mathcal{C}^{(r)}} \tilde{\lambda}_j^{(r)}, \quad \forall c\in\mathcal{C}^{(r)}$\;

    \tcp{Select the top $m$ clients and aggregate}
    $\mathcal{S}^{(r)} \leftarrow \textsc{TopM}(\{\lambda_c^{(r)}\}_{c\in\mathcal{C}^{(r)}},m)$\;
    
    $\lambda_c^{(r)} \leftarrow \lambda_c^{(r)} \big/ \sum_{j\in\mathcal{S}^{(r)}} \lambda_j^{(r)}, \quad \forall c\in\mathcal{S}^{(r)}$\;
    
    $\theta^{(r+1)} \leftarrow \sum_{c\in\mathcal{S}^{(r)}} \lambda_c^{(r)} \theta_c^{(r)}$\;

    \tcp{Learner evaluation}
    Evaluate $\theta^{(r+1)}$ on $\mathcal{S}_{0}$\;
}

\Return $\theta^{(T)}$\;
\end{algorithm}

%% file: 05_experiment.tex
\section{Evaluation Results}
\label{sec:evaluation}
We implement DIC-KT (\emph{Ours}) and representative baselines (see Sec.\ref{sec:baseline}) in  Flower Federated Framework~\cite{beutel2020flower} to systematically evaluate learner-centric knowledge transfer in federated learning. 
All methods follow a common federated training protocol, where at each communication round, a subset of clients is selected to perform local optimization and the resulting updates are transmitted to a central learner for aggregation.

\subsection{Model and Training Settings}

We adopt a standard image classification setting on the CIFAR-10 dataset, which consists of 60,000 color images across 10 classes, with 50,000 training and 10,000 test samples. The data can be partitioned across clients in either homogeneous (IID) or heterogeneous (non-IID) manners to reflect realistic federated learning scenarios. 
We use a convolutional neural network (CNN) consisting of three convolutional blocks followed by a two-layer fully connected classifier head.
All models are trained using stochastic gradient descent (SGD) with momentum.
Each selected client performs $E$ local epochs with a mini-batch size of $64$, momentum of $0.9$, and weight decay of $5\times10^{-4}$.
During each communication round, the learner aggregates client updates (e.g., via weighted averaging) and evaluates model performance after each aggregation step.

\subsection{Federated Data Partition and Learner Setup}
\label{sec: FLsetup}

We simulate a federation consisting of one designated \emph{learner} and $20$ auxiliary clients. 
The CIFAR-10 training set is partitioned across clients under a non-IID regime using a Dirichlet partition. Specifically, each client's class proportions are drawn from a Dirichlet distribution with concentration parameter $\alpha$.
Smaller values of $\alpha$ correspond to more severe label skew.
Unless otherwise stated, we set $\alpha = 0.1$ to induce strong data heterogeneity.

The learner is associated with a fixed target evaluation set of 2{,}000 samples drawn from a Dirichlet distribution with the same concentration parameter (i.e., $\alpha = 0.1$).
This target set defines the learner's objective distribution and remains fixed across all methods and communication rounds, ensuring a consistent basis for comparison.
For methods that require learner-side feedback, we additionally provide access to a small learner proxy dataset with the same label support as the evaluation set. This dataset offers limited supervision for estimating client utility.

\subsection{Baselines and  Accuracy Comparisons}
\label{sec:baseline}

\begin{table*}[t]
\centering
\caption{
Learner test accuracy (\%, mean $\pm$ std over the last 10 rounds).
We evaluate (A) label-exclusive client selection under different numbers,
(B) Dirichlet heterogeneity levels, and
(C) selection ratio under $\alpha{=}0.1$.
Best results in each row are shown in \textbf{bold}. The selection ratio is denoted as $s$ and the Dirichlet concentration is denoted as $\alpha$.
}
\label{tab:all_exp}
\footnotesize
\setlength{\tabcolsep}{4pt} 
\begin{tabular}{l l l l l l l l}
\toprule
Ratio $s$  &  $\alpha$ &
FedAvg  & KMeans & FL+HC & H-Ensemble & Power-of-Choice & DIC-KT(Ours) \\
\midrule

\multicolumn{8}{l}{\textbf{(A) Label-exclusive case}} \\
\midrule
$s = 0.15$ & $\alpha{=}0.1$ &19.91 (± 11.0) & 12.40 (± 12.4) & 78.67 (± 0.34) & 79.31 (± 1.32)  & 61.67 (± 13.86)& \textbf{79.52 (± 1.0)} \\
$s = 0.2$ & $\alpha{=}0.1$ &14.76 (± 12.1) & 41.20 (± 30.3) & \textbf{80.12 (± 0.42)} & 79.31 (± 1.07)  & 62.76 (± 15.67)& 79.32 (± 0.62) \\
$s = 0.35$ & $\alpha{=}0.1$ &42.72 (± 30.9) & 37.14 (± 34.2) & 65.67 (± 0.34) & 66.31 (± 2.62)  & 65.41 (± 11.89)& \textbf{66.78 (± 8.38)} \\
$s = 0.5$ & $\alpha{=}0.1$ & 46.70 (± 31.0) & 47.73 (± 28.9) & 54.67 (± 0.54)& 51.31 (± 8.12)  & 51.67 (±16.67)& \textbf{56.05 (± 22.0)} \\

\midrule
\multicolumn{8}{l}{\textbf{(B) Mismatch levels under the fixed selection ratio}} \\
\midrule
  $s = 0.25$ &$\alpha{=}0.1$& 68.29 (± 13.50) & 67.56 (± 1.12) & 80.58 (± 5.55) & 86.36 (± 1.23)  & 52.04 (± 17.47)& \textbf{92.80 (± 0.45)} \\
 $s = 0.25$ & $\alpha{=}0.5$& 83.32 (± 2.95) & 68.72 (± 0.97) & 83.42 (± 0.65) & 85.36 (± 1.61)  & 79.77 (± 3.27)& \textbf{85.39 (± 1.10)} \\
$s = 0.25$& $\alpha{=}2.0$  & 85.14 (± 1.04) & 77.99 (± 0.97) & 83.21 (± 1.46) & \textbf{86.77 (± 1.61)}  & 84.49 (± 0.99)& 85.51 (± 0.50) \\

\midrule
\multicolumn{8}{l}{\textbf{(C) Different selection ratios under the same mismatch level}} \\
\midrule
$s = 0.15$  & $\alpha{=}0.1$ & 39.97 (± 25.60) & 67.73 (± 17.76) & 84.67 (± 8.12)  & 86.36 (± 1.23) & 49.04 (± 18.44)& \textbf{88.85 (± 0.85)}\\
$s = 0.25$  & $\alpha{=}0.1$ & 56.46 (± 21.50) & 62.27 (± 16.75) & 80.58 (± 5.55) & 86.36 (± 1.01) & 52.04 (± 17.47) & \textbf{90.27 (± 2.83)}\\
$s = 0.3$  & $\alpha{=}0.1$ & 55.93 (± 22.36) & 62.68 (± 22.58) & 81.23 (± 4.35) & 87.75 (± 1.10)  & 56.04 (± 19.12)& \textbf{89.97 (± 0.62)}\\
$s = 0.5$ & $\alpha{=}0.1$ & 68.56 (± 13.02) & 80.01 (± 7.87) & 82.67 (± 7.45) & 87.36 (± 1.09)  & 71.04 (± 11.15)& \textbf{90.52 (± 1.20)} \\

\bottomrule
\end{tabular}
\end{table*}

We compare the proposed learner-centric influence-aware selection method against representative baselines that reflect distinct client selection principles in federated and multi-source transfer learning.
These include uniform random sampling (\textbf{FedAvg})\cite{mcmahan2017fedavg}, clustering-based similarity selection (\textbf{KMeans}), distributional matching via Jensen-Shannon divergence (\textbf{FL+HC})\cite{briggs2020federated}, ensemble-based multi-source transfer (\textbf{H-Ensemble})\cite{wu2024h}, and dynamic high-loss selection (\textbf{Power-of-Choice})\cite{alistarh2020power}.
All methods operate under an identical federated training protocol and differ only in how clients are selected or weighted for aggregation.
Collectively, these baselines span a spectrum from static to dynamic strategies and exhibit varying sensitivity to label coverage, distribution mismatch, and learner-side data scarcity.


 
\begin{figure}[t]
    \centering
    \includegraphics[width=0.8\columnwidth]{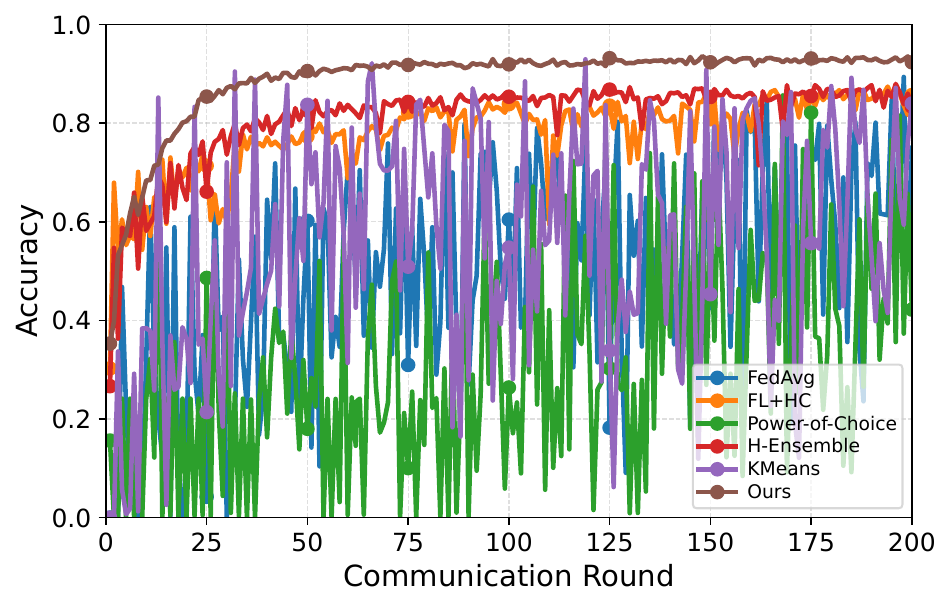}
    \caption{Our influence-aware selection yields consistently higher accuracy and more stable convergence than FedAvg, FL+HC, KMeans, H-Ensemble, and Power-of-Choice.}
    \label{fig:overall}
\end{figure}

\begin{figure}[t]
\centering
\includegraphics[width=0.8\linewidth]{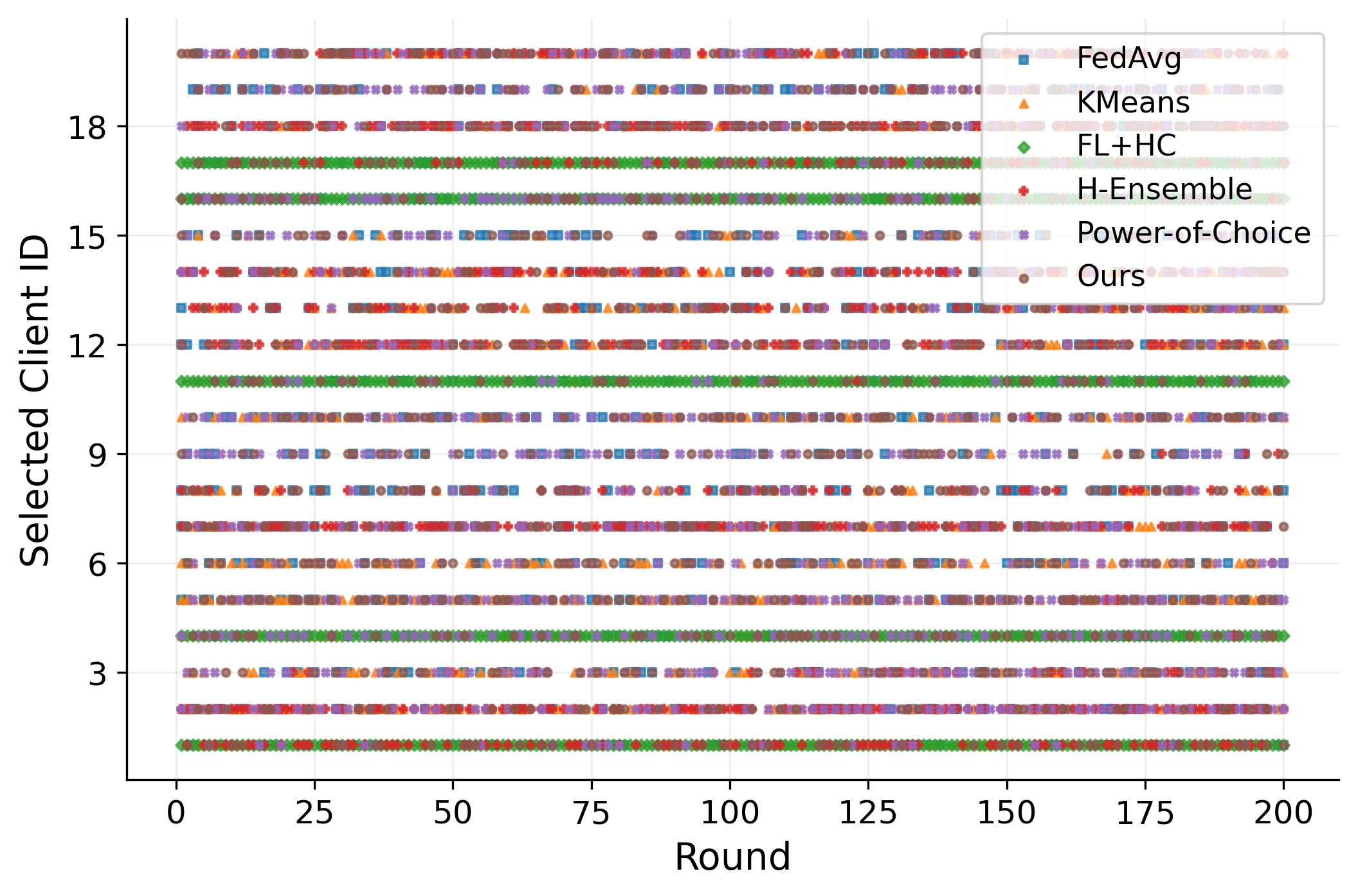}
\caption{Selected client IDs across communication rounds for different client selection strategies. Each marker represents a selected client at a given round.}
\label{fig:selection_ids}
\end{figure}

Fig.~\ref{fig:overall} presents the average learner test accuracies over every 25 communication rounds for DIC-KT and all baseline methods. In this experiment, the Dirichlet concentration $\alpha$ is set to $0.1$ and the client selection ratio $s$ is fixed at 0.25 (i.e., 5 clients are selected per round). Detailed descriptions of the experimental setup can be found in Sec.~\ref{sec:baseline}, while the label distributions for the learner and all 20 clients are illustrated in Fig.~\ref{fig:label_dist_all_cases}-(d).
To further investigate the observed performance differences, Fig.~\ref{fig:selection_ids} visualizes the client selection patterns (i.e., the selected client IDs) across rounds under DIC-KT and all baselines.

Methods such as FedAvg (random), KMeans, and Power-of-Choice exhibit either highly random or overly repetitive selection behaviors. As a result, their performance remains unstable, as they lack an explicit mechanism to estimate learner-centric utility. In contrast, H-Ensemble and FL+HC partially account for distributional alignment, obtaining improved stability and higher accuracy.
Our proposed influence-aware method (DIC-KT) consistently achieves the highest learner test accuracy and shows a more stable convergence trajectory across communication rounds. This indicates that our method identifies and prioritizes clients that contribute positively to the learner's objective, enabling more effective knowledge transfer compared to static similarity matching or loss-driven heuristics.

\subsection{Sensitivity Analysis and Discussion}
We further evaluate learner-centric knowledge transfer performance across four complementary sensitivity dimensions:
(a) label exclusivity,
(b) degree of distribution mismatch, 
(c) client selection ratio, and
(d) learner proxy set size.

\begin{figure}[t]
\centering
\subfloat[Client label distribution structure.]{%
    \includegraphics[width=0.45\linewidth,height=4cm,keepaspectratio]{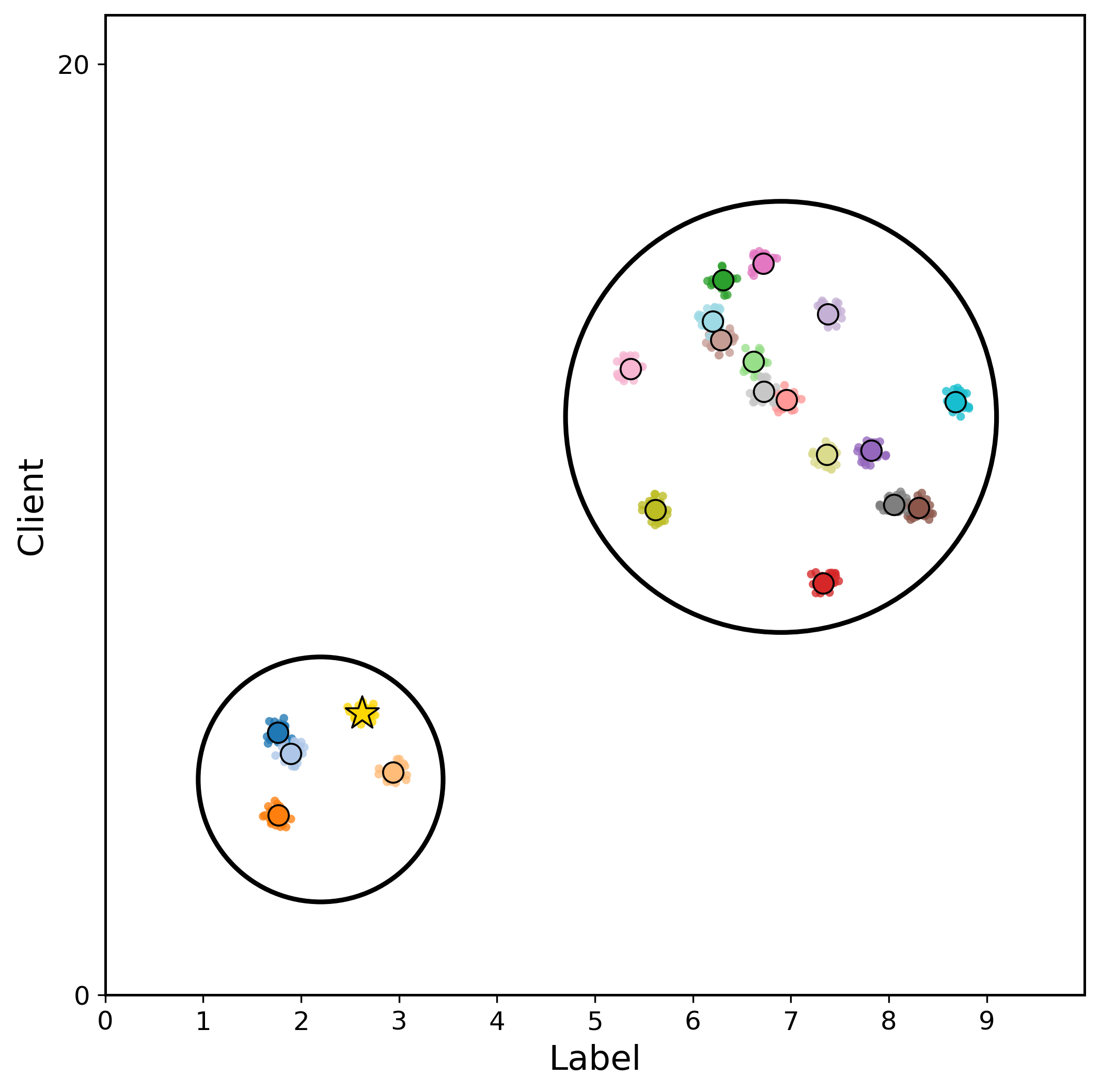}
}
\hfill
\subfloat[Exclusive-case JSD distance between each client and the learner distribution.]{%
    \includegraphics[width=0.48\linewidth,height=4.4cm,keepaspectratio]{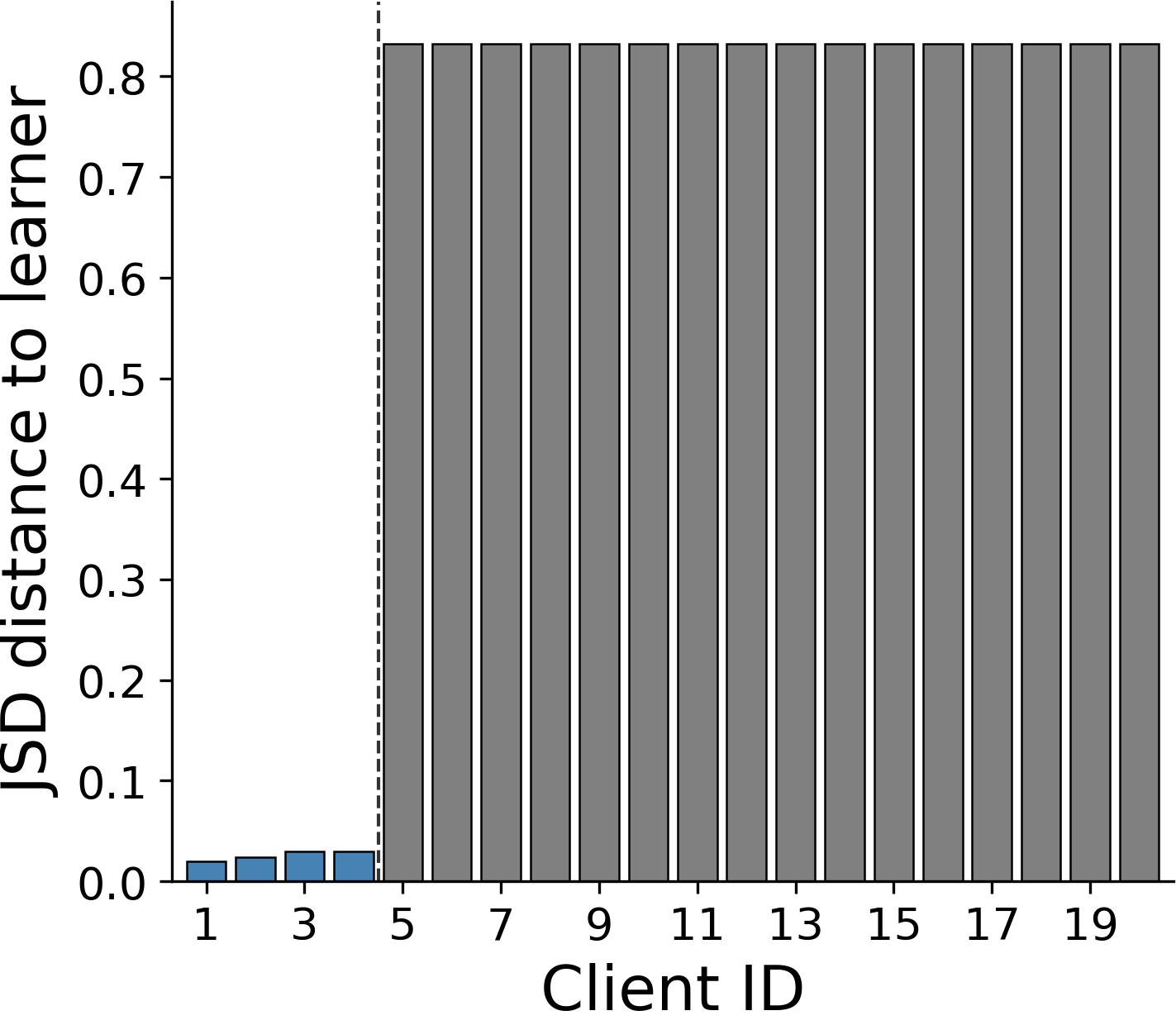}
}
\caption{Illustration of the label-exclusive setting and the corresponding distribution distance. (a) The left plot shows that the learner (yellow star) and clients containing labels $\{0,1,2,3\}$ form the left cluster, while the remaining clients containing labels $\{4,5,6,7,8,9\}$ form the right cluster. (b) The right plot shows the JSD distance between each client distribution and the learner target distribution.}
\label{fig:label_case_distribution1}
\end{figure}

\paragraph{Label-exclusive Case}
We first consider a manually constructed label-exclusive setting to explicitly control label alignment between the learner and clients.
Specifically, as illustrated in Fig.~\ref{fig:label_case_distribution1}(a), a small subset of clients (i.e., clients 1,2,3,4 for selection ratio $s = 0.2$) is assigned to share the same label support as the learner's target data and thus lies in the same distributional cluster with the learner, while the remaining clients collectively cover the other CIFAR-10 classes. Fig.~\ref{fig:label_case_distribution1}(b) further shows the JSD distance between each client distribution and the learner target distribution, where the four label-aligned clients exhibit notably smaller divergence values. 
This design isolates the effect of label-aligned knowledge transfer and creates a challenging scenario in which only a few clients are directly relevant to the learner's objective.

As shown in Table~\ref{tab:all_exp}-(A), in this case, random and clustering-based selection methods exhibit pronounced instability and high variance, as they frequently select label-mismatched clients that provide limited or even misleading supervision.
Similarity-based methods, such as FL+HC and H-Ensemble, improve robustness by repeatedly favoring distributionally aligned sources, but their performance remains sensitive to the client selection ratio and lacks adaptivity once the relevant clients are identified.
In contrast, the proposed influence-aware method consistently achieves the highest accuracy across all selection ratios, indicating its ability to reliably identify, prioritize, and exploit the small subset of label-aligned clients, while suppressing negative transfer from irrelevant sources.

\paragraph{Different Mismatch Levels}
In Table~\ref{tab:all_exp}-(B), we fix the client selection ratio ($s=0.25$ clients) while increasing the Dirichlet concentration parameter from $\alpha=0.1$ to $\alpha=2.0$. As a result, learner-client distribution mismatch is progressively reduced, and performance gaps across all methods narrow.
Under severe heterogeneity ($\alpha=0.1$), our approach achieves the highest accuracy, improving over FedAvg and KMeans by approximately $35.9\%$ and $37.4\%$, respectively, and outperforming the difficulty-based Power-of-Choice method by about $78.4\%$.
Compared with the static and ensemble baselines, our method improves by about $15.2\%$ over FL+HC and $7.5\%$ over H-Ensemble, while also exhibiting substantially lower variance, demonstrating strong robustness under extreme distribution mismatch.

Such a performance gap narrows as heterogeneity decreases ($\alpha=0.5$). The influence-aware selection remains marginally better, achieving roughly a $2.4\%$ improvement over FL+HC and performing on par with H-Ensemble.
In the near-uniform case ($\alpha=2.0$), where client and learner distributions are well aligned, differences further diminish, with our method remaining competitive and within $1.5\%$ of the best-performing baseline.
Overall, these results indicate that the proposed influence-aware method delivers substantial relative accuracy gains under severe mismatch, while naturally converging toward comparable performance as distributional alignment improves.

\paragraph{Selection Ratio Sensitivity}
In Table~\ref{tab:all_exp}-(C), we conduct a sensitivity analysis on the selection ratio by varying the number of clients selected per round. The results reveal that increased participation alone does not necessarily improve learner performance.
Random and difficulty-based selection exhibit large performance fluctuations as the ratio changes, highlighting inefficiencies in exploring the client space.
On the other hand, similarity-based methods benefit from moderate selection ratios but saturate quickly due to static selection signals.
In contrast, our influence-aware selection maintains consistently strong performance across all ratios, suggesting that the dynamic prioritization of informative clients is more effective than the indiscriminate scaling of participation.

\begin{figure}[t]
\centering
\includegraphics[width=0.7\linewidth]{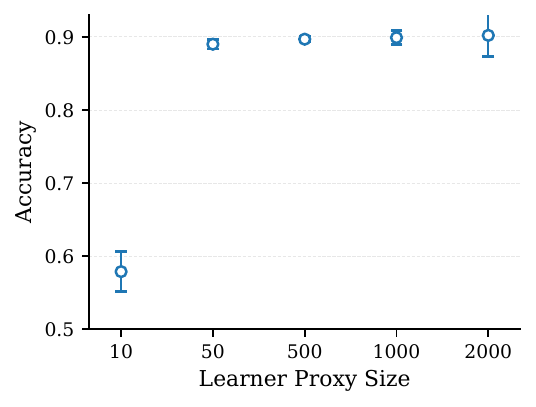}
\caption{Effect of the learner proxy set size on knowledge transfer performance. 
The plot shows the average accuracy over the last ten rounds, with error bars indicating the standard deviation. 
Even a small learner proxy set provides sufficient guidance for effective knowledge transfer.}
\label{fig:ps_size_transfer}
\end{figure}

\paragraph{Learner Proxy Size Sensitivity}
We further examine how the size of the learner proxy dataset affects knowledge transfer performance by evaluating five different set sizes.
As shown in Fig.~\ref{fig:ps_size_transfer}, learner accuracy is computed as the average over the final ten training rounds, with error bars representing the standard deviation across rounds. 
%
The results reveal a clear threshold effect. When the proxy set is extremely small (e.g., 10 samples), the learner experiences a substantial drop in accuracy, indicating that the limited data is insufficient to reliably estimate client utility and guide effective knowledge transfer. 
However, once the proxy size increases to a modest level (e.g., 50 samples), performance improves dramatically and quickly approaches that of much larger proxy sets (e.g., 500, 1000, or 2000 samples). 
Beyond this point, further increasing the proxy size yields only marginal gains, suggesting that the proposed influence-aware selection can already obtain sufficiently accurate estimates of client contribution using a relatively small amount of learner data. 


Overall, the above results demonstrate that learner-centric federated learning is highly sensitive to label coverage, distribution mismatch, and client participation ratio.
Static similarity metrics and difficulty-based heuristics are insufficient to reliably support effective knowledge transfer.
By dynamically estimating each client's marginal contribution to the learner's objective, the proposed influence-aware method enables adaptive exploration of the client space and achieves robust, stable, and consistently superior transfer performance across all evaluated settings.




%% file: 06_conclusion.tex
\section{Conclusion}
\label{sec:conclusion}
In this paper, we study federated learning from a learner-centric perspective and show that under learner-client distribution mismatch, effective collaboration is fundamentally a problem of \emph{knowledge transfer} rather than uniform population-level optimization.
To address this challenge, our proposed Dynamic Influence-Aware Control for Knowledge Transfer (DIC-KT) method dynamically estimates each client's marginal contribution to the learner’s objective using a small target proxy set and jointly optimizes client selection and aggregation.
Experiments on CIFAR-10 demonstrate that DIC-KT consistently outperforms both static and dynamic baselines, achieving higher accuracy with lower variance, particularly under severe distribution mismatch.  In the future, we plan to extend DIC-KT to real-world applications such as autonomous driving systems, where vehicles collect highly heterogeneous driving-scene data across diverse environments. We will study how influence-aware client selection improves knowledge transfer under severe distribution misalignment, enabling more reliable and efficient model adaptation in safety-critical scenarios.

%% file: IEEEcustom.bib
@STRING{PMLR= "Proceedings of Machine Learning Research"}

@article{Zhang2025AHS,
  title={A High-Dimensional Statistical Method for Optimizing Transfer Quantities in Multi-Source Transfer Learning},
  author={Qingyue Zhang and Haohao Fu and Guanbo Huang and Yaoyuan Liang and Chang Chu and Tianren Peng and Yanru Wu and Qi Li and Yang Li and Shao-Lun Huang},
  journal={Advances in Neural Information Processing Systems 38},
  year={2025}

}

@inproceedings{wang2020convergence,
  title={On the Convergence of FedAvg on Non-IID Data},
  author={Wang, Jianyu and Shen, Chengchao and Ding, Guiguang and Chen, Liang},
  booktitle={Advances in Neural Information Processing Systems (NeurIPS)},
  year={2020}
}

@article{kairouz2021advances,
  title={Advances and open problems in federated learning},
  author={Kairouz, Peter and McMahan, H Brendan},
  journal={Foundations and trends in machine learning},
  volume={14},
  number={1-2},
  pages={1--210},
  year={2021},
  publisher={Emerald Publishing Limited}
}

@inproceedings{briggs2020federated,
  title={Federated learning with hierarchical clustering of local updates to improve training on non-IID data},
  author={Briggs, Christopher and Fan, Zhong and Andras, Peter},
  booktitle={2020 international joint conference on neural networks (IJCNN)},
  pages={1--9},
  year={2020},
  organization={IEEE}
}

@inproceedings{mcmahan2017fedavg,
  title     = {Communication-Efficient Learning of Deep Networks from Decentralized Data},
  author    = {McMahan, Brendan and Moore, Eider and Ramage, Daniel and Hampson, Seth and y Arcas, Blaise Aguera},
  booktitle = {Proceedings of the 20th International Conference on Artificial Intelligence and Statistics (AISTATS)},
  year      = {2017}
}

@inproceedings{li2020fedprox,
  title     = {Federated Optimization in Heterogeneous Networks},
  author    = {Li, Tian and Sahu, Anit Kumar and Talwalkar, Ameet and Smith, Virginia},
  booktitle = {Proceedings of MLSys},
  year      = {2020}
}

@inproceedings{karimireddy2020scaffold,
  title     = {SCAFFOLD: Stochastic Controlled Averaging for Federated Learning},
  author    = {Karimireddy, Sai Praneeth and others},
  booktitle = {Proceedings of ICML},
  year      = {2020}
}

@inproceedings{zhao2018federated,
  title={Federated learning with non-IID data},
  author={Zhao, Yue and Li, Meng and Lai, Liangzhen and Suda, Naveen and Civin, David and Chandra, Vikas},
  booktitle={NeurIPS Workshop on Federated Learning},
  year={2018}
}

@inproceedings{li2021ditto,
  title={Ditto: Fair and robust federated learning through personalization},
  author={Li, Tian and Hu, Sheng and Beirami, Ahmad and Smith, Virginia},
  booktitle={Proceedings of the 38th International Conference on Machine Learning (ICML)},
  year={2021}
}

@inproceedings{fallah2020personalized,
  title={Personalized federated learning with theoretical guarantees},
  author={Fallah, Alireza and Mokhtari, Aryan and Ozdaglar, Asuman},
  booktitle={Advances in Neural Information Processing Systems (NeurIPS)},
  year={2020}
}

@inproceedings{smith2017federated,
  title={Federated multi-task learning},
  author={Smith, Virginia and Chiang, Chao-Kai and Sanjabi, Maziar and Talwalkar, Ameet},
  booktitle={Advances in Neural Information Processing Systems (NeurIPS)},
  year={2017}
}

@article{li2021survey,
  title={Personalized federated learning: A survey},
  author={Li, Qinbin and Diao, Yiqun and Chen, Quan and He, Bingsheng},
  journal={ACM Computing Surveys},
  year={2023}
}

@inproceedings{bonawitz2019towards,
  title     = {Towards Federated Learning at Scale: System Design},
  author    = {Bonawitz, Keith and Eichner, Hubert and Grieskamp, Wolfgang and others},
  booktitle = {Proceedings of MLSys},
  year      = {2019}
}

@inproceedings{Xue2020TowardUT,
  title={Toward understanding the influence of individual clients in federated learning},
  author={Xue, Yihao and Niu, Chaoyue and Zheng, Zhenzhe and Tang, Shaojie and Lyu, Chengfei and Wu, Fan and Chen, Guihai},
  booktitle={Proceedings of the AAAI Conference on Artificial Intelligence},
  volume={35},
  number={12},
  pages={10560--10567},
  year={2021}
}

@inproceedings{alistarh2020power,
  title={Towards understanding biased client selection in federated learning},
  author={Cho, Yae Jee and Wang, Jianyu and Joshi, Gauri},
  booktitle={International conference on artificial intelligence and statistics},
  pages={10351--10375},
  year={2022},
  organization={PMLR}
}

@inproceedings{you2021logme,
  title     = {LogME: Practical Assessment of Pre-trained Models for Transfer Learning},
  author    = {You, Kaichao and others},
  booktitle = {ICML},
  year      = {2021}
}

@inproceedings{liu2021model,
  title     = {Model-based Weighting for Transfer Learning},
  author    = {Liu, Yang and others},
  booktitle = {NeurIPS},
  year      = {2021}
}

@article{pan2010survey,
  title={A Survey on Transfer Learning},
  author={Pan, Sinno Jialin and Yang, Qiang},
  journal={IEEE Transactions on Knowledge and Data Engineering},
  year={2010}
}

@techreport{Krizhevsky09learningmultiple,
  author = {Alex Krizhevsky},
  title = {Learning multiple layers of features from tiny images},
  institution = {University of Toronto},
  year = {2009},
  type = {Technical Report}
}

@article{ben2010theory,
  title={A theory of learning from different domains},
  author={Ben-David, Shai and Blitzer, John and Crammer, Koby and Pereira, Fernando},
  journal={Machine Learning},
  volume={79},
  number={1-2},
  pages={151--175},
  year={2010}
}

@inproceedings{mansour2009domain,
  title={Domain adaptation with multiple sources},
  author={Mansour, Yishay and Mohri, Mehryar and Rostamizadeh, Afshin},
  booktitle={NeurIPS},
  year={2009}
}

@inproceedings{gong2013connecting,
  title={Connecting the dots with landmarks: Discriminatively learning domain-invariant features},
  author={Gong, Boqing and Shi, Yelong and Sha, Fei and Grauman, Kristen},
  booktitle={ICML},
  year={2013}
}

@inproceedings{zhao2018adversarial,
  title={Adversarial multiple source domain adaptation},
  author={Zhao, Han and Zhang, Shanghang and Wu, Guanhang and Moura, Jose and Costeira, Joao and Gordon, Geoffrey},
  booktitle={NeurIPS},
  year={2018}
}

@inproceedings{peng2019moment,
  title={Moment matching for multi-source domain adaptation},
  author={Peng, Xingchao and Bai, Qiong and Xia, Xide and Huang, Zijun and Saenko, Kate and Wang, Bo},
  booktitle={ICCV},
  year={2019}
}

@inproceedings{guo2018selective,
  title={Selective transfer learning for cross domain recommendation},
  author={Guo, Guibing and Zhang, Jie and Yorke-Smith, Neil},
  booktitle={AAAI},
  year={2018}
}

@ARTICLE{9084352,
  author={Li, Tian and Sahu, Anit Kumar and Talwalkar, Ameet and Smith, Virginia},
  journal={IEEE Signal Processing Magazine}, 
  title={Federated Learning: Challenges, Methods, and Future Directions}, 
  year={2020},
  volume={37},
  number={3},
  pages={50-60},
  doi={10.1109/MSP.2020.2975749}}

@article{tong2021mathematical,
  title={A mathematical framework for quantifying transferability in multi-source transfer learning},
  author={Tong, Xinyi and Xu, Xiangxiang and Huang, Shao-Lun and Zheng, Lizhong},
  journal={Advances in Neural Information Processing Systems},
  volume={34},
  pages={26103--26116},
  year={2021}
}

@inproceedings{wu2024h,
  title={H-ensemble: An information theoretic approach to reliable few-shot multi-source-free transfer},
  author={Wu, Yanru and Wang, Jianning and Wang, Weida and Li, Yang},
  booktitle={Proceedings of the AAAI Conference on Artificial Intelligence},
  volume={38},
  number={14},
  pages={15970--15978},
  year={2024}
}

@article{sun2011two,
  title={A two-stage weighting framework for multi-source domain adaptation},
  author={Sun, Qian and Chattopadhyay, Rita and Panchanathan, Sethuraman and Ye, Jieping},
  journal={Advances in neural information processing systems},
  volume={24},
  year={2011}
}

@inproceedings{wang2020optimizing,
  title={Optimizing federated learning on non-iid data with reinforcement learning},
  author={Wang, Hao and Kaplan, Zakhary and Niu, Di and Li, Baochun},
  booktitle={IEEE INFOCOM 2020-IEEE conference on computer communications},
  pages={1698--1707},
  year={2020},
  organization={IEEE}
}

@inproceedings{chai2020tifl,
  title={TiFL: A Tier-based Federated Learning System},
  author={Chai, Zheng and Ali, Ahsan and Zawad, Syed and Anwar, Ahsan and Zhou, Yi and Baracaldo, Nathalie and Ludwig, Heiko and Yan, Feng},
  booktitle={Proceedings of the 29th International Symposium on High-Performance Parallel and Distributed Computing (HPDC)},
  year={2020},
  pages={125--136}
}

@inproceedings {273723,
author = {Fan Lai and Xiangfeng Zhu and Harsha V. Madhyastha and Mosharaf Chowdhury},
title = {Oort: Efficient Federated Learning via Guided Participant Selection},
booktitle = {15th {USENIX} Symposium on Operating Systems Design and Implementation ({OSDI} 21)},
year = {2021},
isbn = {978-1-939133-22-9},
pages = {19--35},
publisher = {{USENIX} Association},
month = jul
}

@INPROCEEDINGS{8761315,
  author={Nishio, Takayuki and Yonetani, Ryo},
  booktitle={ICC 2019 - 2019 IEEE International Conference on Communications (ICC)}, 
  title={Client Selection for Federated Learning with Heterogeneous Resources in Mobile Edge}, 
  year={2019},
  volume={},
  number={},
  pages={1-7},
  doi={10.1109/ICC.2019.8761315}}

@article{Vehtari2015PracticalBM,
  title={Practical Bayesian model evaluation using leave-one-out cross-validation and WAIC},
  author={Vehtari, Aki and Gelman, Andrew and Gabry, Jonah},
  journal={Statistics and computing},
  volume={27},
  number={5},
  pages={1413--1432},
  year={2017},
  publisher={Springer}
}

@article{beutel2020flower,
  title={Flower: A Friendly Federated Learning Research Framework},
  author={Beutel, Daniel J and Topal, Taner and Mathur, Akhil and Qiu, Xinchi and Fernandez-Marques, Javier and Gao, Yan and Sani, Lorenzo and Kwing, Hei Li and Parcollet, Titouan and Gusmão, Pedro PB de and Lane, Nicholas D},
  journal={arXiv preprint arXiv:2007.14390},
  year={2020}
}

@incollection{wang2020principled,
  title={A principled approach to data valuation for federated learning},
  author={Wang, Tianhao and Rausch, Johannes and Zhang, Ce and Jia, Ruoxi and Song, Dawn},
  booktitle={Federated Learning: Privacy and Incentive},
  pages={153--167},
  year={2020},
  publisher={Springer}
}

@inproceedings{xie2019zeno,
  title={Zeno: Distributed Stochastic Gradient Descent with Suspicion-based Fault-tolerance},
  author={Xie, Cong and Koyejo, Oluwasanmi and Gupta, Indranil},
  booktitle={International Conference on Machine Learning},
  year={2019}
}
